\documentclass[journal]{IEEEtran}

\usepackage{graphicx}
\usepackage[caption=false, font=footnotesize]{subfig}

\usepackage{fancyhdr}

\usepackage{xcolor}         % use text color
\usepackage{booktabs}       % professional-quality tables
\usepackage{array}
\usepackage{tikz}

\newcommand\Tstrut{\rule{0pt}{2.6ex}}         % = `top' strut
\newcommand\Bstrut{\rule[-0.9ex]{0pt}{0pt}}   % = `bottom' strut
\newcommand\crule[3][black]{\textcolor{#1}{\rule{#2}{#3}}}

\usepackage{makecell}       % multi-line in table
\usepackage[hyphens]{url}   % for url line breaks
\usepackage{hyperref}       % for hyperlinks
\usepackage{amsfonts,amsmath,amssymb,amsthm}       % blackboard math symbols
\usepackage{nicefrac}       % compact symbols for 1/2, etc.
\usepackage{algorithm, algpseudocode} % nice algorithm 
\usepackage{cleveref}       %automatically detect the ref type

\usepackage{bm}             % bold math
\usepackage[justification=centering]{caption} 
\usepackage{stfloats}
\begin{document}

\title{Decision Making in Monopoly using a \\ Hybrid Deep Reinforcement Learning Approach}
% note positions of commas and non-breaking spaces ( ~ ) LaTeX will not break
\author{
	Trevor Bonjour$^{1*}$, Marina Haliem$^{1*}$, Aala Alsalem$^{1}$, Shilpa Thomas$^{2}$, Hongyu Li$^{2}$, \\ Vaneet Aggarwal$^{1}$, Mayank Kejriwal$^{2}$, and Bharat Bhargava$^{1}$\thanks{
	© 2022 IEEE.  Personal use of this material is permitted.  Permission from IEEE must be obtained for all other uses, in any current or future media, including reprinting/republishing this material for advertising or promotional purposes, creating new collective works, for resale or redistribution to servers or lists, or reuse of any copyrighted component of this work in other works.
	
	$^*$ Equal contribution 
	
	$^1$ Purdue University, $^2$ University of Southern California	}
	}

\maketitle
\begin{abstract}
Learning to adapt and make real-time informed decisions in a dynamic and complex environment is a challenging problem. Monopoly is a popular strategic board game that requires players to make multiple decisions during the game. Decision-making in Monopoly involves many real-world elements such as strategizing, luck, and modeling of opponent's policies. In this paper, we present novel representations for the state and action space for the full version of Monopoly and define an improved reward function. Using these, we show that our deep reinforcement learning agent can learn winning strategies for Monopoly against different fixed-policy agents. In Monopoly, players can take multiple actions even if it is not their turn to roll the dice. Some of these actions occur more frequently than others, resulting in a skewed distribution that adversely affects the performance of the learning agent. To tackle the non-uniform distribution of actions, we propose a hybrid approach that combines deep reinforcement learning (for frequent but complex decisions) with a fixed-policy approach (for infrequent but straightforward decisions). We develop learning agents using proximal policy optimization (PPO) and double deep Q-learning (DDQN) algorithms and compare the standard approach to our proposed hybrid approach. Experimental results show that our hybrid agents outperform standard agents by 20\% in the number of games won against fixed-policy agents. The hybrid PPO agent performs the best with a win rate of 91\% against fixed-policy agents.
\end{abstract}
\begin{IEEEkeywords}
Monopoly, Deep Reinforcement Learning, Decision Making, Proximal Policy Optimization, Double Deep Q-Learning.
\end{IEEEkeywords}
\section{Introduction}
% \IEEEPARstart{T}{his} demo file is intended to \cite{bailis} serve as a ``starter file''
% for IEEE journal papers produced under \LaTeX\ using
% IEEEtran.cls version 1.8b and later.
% % You must have at least 2 lines in the paragraph with the drop letter
% % (should never be an issue)
% I wish you the best of success.

\IEEEPARstart{D}{espite} numerous advances in deep reinforcement learning (DRL), the majority of successes have been in two-player, zero-sum games, where it is guaranteed to converge to an optimal policy \cite{zerosum}, such as Chess and Go \cite{silver2016mastering}. Rare (and relatively recent) exceptions include Blade \& Soul \cite{oh2021creating}, no-press diplomacy \cite{diplo1}, Poker\footnote{We note that, even in this case, a two-player version of Texas Hold 'em was initially assumed \cite{moravvcik2017deepstack} but later superseded by a multi-player system.} \cite{multiplayerpoker},  and StarCraft \cite{starcraft}, \cite{star2}. In particular, there has been little work on agent development for the full 4-player game of Monopoly, despite it being one of the most popular strategic board games in the last 85 years.

Monopoly is a turn-based real-estate game in which the goal is to remain financially solvent. The objective is to force the opponents into bankruptcy by buying, selling, trading, and improving (building a house or a hotel) pieces of property. A player is only allowed to improve property when they achieve a monopoly. A monopoly is when a player owns all the properties that are part of the same color group. The game resembles the real-life business practice of cornering the market to achieve a real-estate monopoly.

During the game, a player can take multiple actions even when it is not their turn to roll the dice. Imagine you are in the middle of playing Monopoly with friends. It is not your turn to roll the dice, but one of your friends just acquired a property that will give you a monopoly. You know you will need that property if you want to have a chance at winning the game. You initiate a trade request, but you need to make an offer that they will probably accept. You need to think about an amount of money you could offer, or if you have a property that might be of interest to them to offer as an exchange for the property of interest. Maybe you need to mortgage or sell a property to generate cash for the trade - would it even be worth it in the long run. This scenario is a snapshot in time of how many different decisions one needs to make during Monopoly. This complexity makes it a fascinating but challenging problem to tackle.

Previous attempts \cite{bailis, recent} at Monopoly overlook these complexities and consider a simplified version of the game. In both, the authors model Monopoly as a Markov Decision Process (MDP).\cite{bailis} gives a novel representation for the state space. \cite{recent} find that a higher-dimensional representation of the state improves the learning agent's performance. However, both these attempts consider a very limited set of actions: \textit{buy, sell, do nothing} in case of \cite{bailis} and only \textit{buy, do nothing} in case of \cite{recent}. Unlike previous attempts, we do not simplify the action space in Monopoly. Instead, we consider all possible actions (\Cref{table_action_type}), including trades, to make the game as realistic as possible. This consideration makes the task more challenging since we now need to deal with a high-dimensional action space. 

We observe that neither of the earlier state representations contains enough information for the agent to learn winning strategies for Monopoly when considering all the actions. To deal with the high-dimensional action space, we develop an enhanced state space that provides a higher representation power and helps the agent consistently get high win rates against other fixed-policy baseline agents. \cite{recent} use a sparse reward function where the agent receives a reward at the end of each game. Our experiments show that a sparse reward function is not ideal and cannot handle the complexities accompanying the full version of Monopoly. \cite{bailis} use a dense reward function where the agent receives a reward within a game after taking any action. We formulate a dense reward function that performs better than the one given by \cite{bailis}. %Our experiments show that we get the best performance by combining the dense and sparse reward functions. We develop a DRL agent that consistently wins 25\% more games than the best fixed-policy agent.}

In Monopoly, some actions occur more frequently than others resulting in a skewed distribution. For instance, a  player is allowed to \textit{trade}  with other players at any point in the game, but a player can only \textit{buy} an unowned property when they land on the property square. This rare occurrence of a particular state-action pair increases the computational complexity for a standard DRL agent. There is already some evidence emerging that a pure DRL approach may not always be the only (or even best) solution for solving a complex decision-making task. Recently hybrid DRL approaches have surfaced that result in faster convergence, sometimes to a better policy, in other domains such as operations \cite{lee2020algorithm}, robotics \cite{chen2020combining, xiong2020comparison}, and autonomous vehicles \cite{likmeta2020combining, wang2019lane, guo2021hybrid}. To deal with the non-uniform distribution of actions, we propose a hybrid DRL approach for Monopoly. Specifically, we use a fixed-policy approach for infrequent but straightforward decisions and use DRL for frequent but complex decisions. 
We show that our hybrid agents have a faster convergence rate and higher win rates against baseline agents when compared to the standard DRL agents.

We summarize the key contributions of the paper as follows:
\begin{itemize}
    \item We consider all decisions that a player may need to make during Monopoly and develop a novel and comprehensive action space representation (\Cref{sec_action}).
    \item We design an enhanced state space representation (\Cref{sec_state}) and an improved reward function (\Cref{sec_reward}) for Monopoly, using which the learning agents converge sooner and to a better policy in contrast to previous attempts (\Cref{sec_relwork}).
    \item We develop standard DRL-based agents (\Cref{sec_standard}) that learn winning strategies for Monopoly against different fixed-policy agents. The standard DRL agents win at least 25\% more games than the best fixed-policy agent.
    \item We devise a hybrid approach (\Cref{sec_hybrid}) to solve the complex decision-making task using DRL for a subset of decisions in conjunction with fixed-policy for infrequent actions. During training, the hybrid agents converge sooner and to a better policy as compared to the standard DRL agents. Our experiments (\Cref{exp_results}) show that the hybrid agent outperforms the standard learning agent by more than 20\% in the number of games won against the fixed-policy agents.
    \item We develop a complete four-player open-sourced simulator for Monopoly (\Cref{sec_simulator}) together with three different fixed-policy baseline agents. The baseline agents (\Cref{sec_baseline}) are implemented based on common winning strategies used by human players in Monopoly tournaments.
\end{itemize}
\section{Background}

\begin{figure}
\centering
\includegraphics[width=0.40\textwidth]{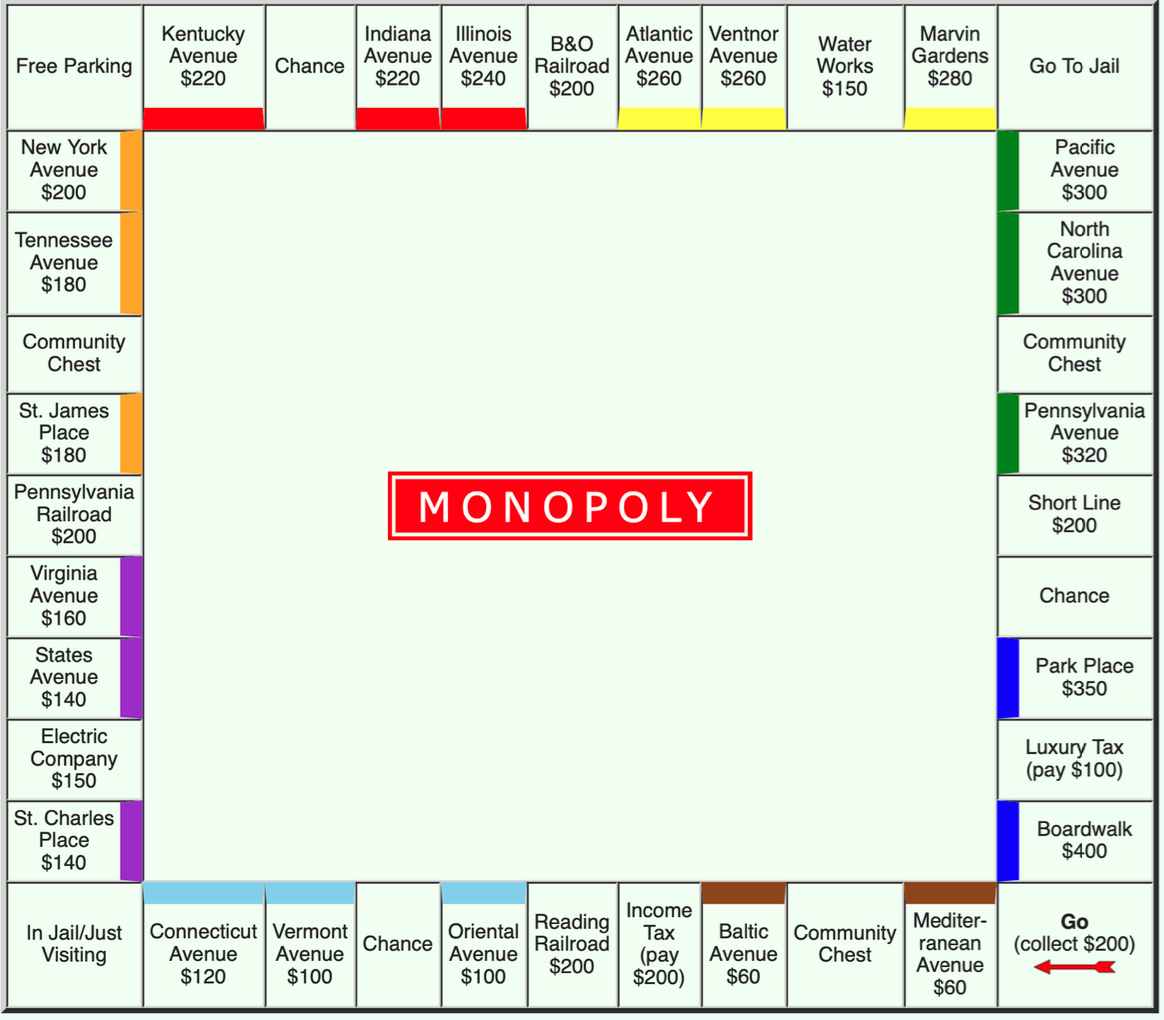}
\caption{Monopoly game board}
\label{fig_board}
\end{figure}

\subsection{Monopoly}

Monopoly is a board game where players take turns rolling a pair of unbiased dice and make decisions based on their position on the board. \Cref{fig_board} shows the conventional Monopoly game board that consists of 40 square locations. These include 28 property locations, distributed among eight color groups (22 real-estate properties), four railroads, and two utility, that players can buy, sell, and trade. Additionally, there are two tax locations that charge players a tax upon landing on them, six card locations that require players to pick a card from either the \textit{community chest} card deck or the \textit{chance} card deck, the \textit{jail} location, the \textit{go to jail} location, the \textit{go} location, and the \textit{free parking} location. Our game schema also specifies all assets, their corresponding purchase prices, rents, and color. Each square shows the purchase prices that correspond to an asset in \Cref{fig_board}. In Monopoly, players act as property owners who seek to buy, sell, improve or trade these properties.  The winner is the one who forces every other player into bankruptcy.

\subsection{Markov Decision Process}
An MDP is defined by the tuple $\langle \mathcal{S}, \mathcal{A}, \mathcal{T}, \mathcal{R} \rangle$ where $\mathcal{S}$ is the set of all possible states and $\mathcal{A}$ is the set of all possible actions. The transition function $\mathcal{T}:\mathcal{S} \times \mathcal{A} \times \mathcal{S} \rightarrow [0, 1]$ is the probability that an action $a\in\mathcal{A}$ in state $s\in\mathcal{S}$ will lead to a transition to state $s'\in\mathcal{S}$. The reward function $\mathcal{R}:\mathcal{S}\times\mathcal{A}\times\mathcal{S} \rightarrow \mathbb{R}$ defines the immediate reward that an agent would receive after executing action $a$ resulting in a transition from state $s$ to $s'$. 

\subsection{Reinforcement Learning}
Solving an MDP yields a policy $\pi : \mathcal{S} \rightarrow \mathcal{A} $, which is a mapping from states to actions. An optimal policy $\pi^*$ maximizes the expected sum of rewards. Reinforcement Learning (RL) is a popular approach to solve an MDP \cite{sutton2018reinforcement} without explicit specification of the transition probabilities. In RL, an agent interacts with the environment in discrete time steps in order to learn the optimal policy through trial and error.
 
% Due to the complexity of the Monopoly environment and the large state and action space it imposes, traditional RL methods like Q-learning \cite{watkins1992q} or REINFORCE \cite{williams1992simple} cannot be directly applied. 
DRL \cite{arulkumaran2017deep} makes use of deep neural networks to approximate the optimal policy or the value function to deal with the limitations of traditional RL methods. The use of deep neural networks as function approximators enables powerful generalization but requires critical decisions about representations. Poor design choices can result in estimates that diverge from the optimal policy \cite{baird1995residual, whiteson2006evolutionary, fujimoto2018addressing}. Existing model-free DRL methods are broadly characterized into \textit{policy gradient} and \textit{value-based} methods.
 
Policy gradient methods use deep networks to optimize the policy directly. Such methods are useful for physical control where the action space is continuous. Some popular policy gradient methods are Deep deterministic policy gradient (DDPG) \cite{lillicrap2015continuous}, asynchronous advantage actor-critic (A3C) \cite{mnih2016asynchronous}, trust region policy optimization (TRPO) \cite{schulman2015trust}, proximal policy optimization (PPO) \cite{schulman2017proximal}.

Value based methods, on the other hand, are based on estimating the value of being in a given state. Deep Q-Network (DQN) \cite{mnih2015human} is a well-known value-based DRL method. There have been many extensions of the DQN algorithm over the past few years, including  double DQN (DDQN) \cite{van2016deep},  distributed DQN \cite{nair2015massively}, prioritized DQN \cite{schaul2015prioritized}, dueling DQN \cite{wang2016dueling}, asynchronous DQN \cite{mnih2016asynchronous}, and rainbow DQN \cite{hessel2018rainbow}. 

In this paper, we implement both a policy gradient method (PPO) and a value based method (DDQN) to train our standard DRL and hybrid agents (\Cref{sec_approach}).

\section{Related Work} \label{sec_relwork}
Despite the popularity of Monopoly, a learning-based approach for decision-making for the full game has not been studied previously. There are older attempts to model Monopoly as a Markov Process such as \cite{old}. \cite{bailis} and more recently \cite{recent} propose modeling Monopoly as an MDP. However, both attempts consider a simplified version of the game. Specifically, both consider a limited set of actions (\textit{buy, sell, do nothing}) with neither work considering trades between players. In \cite{bailis}, an RL agent is trained and tested against a random and a fixed-policy agent.  \cite{bailis} employs a Q-learning strategy along with a neural network. In recent work \cite{recent}, authors apply a feed-forward neural network with the concept of experience replay to learn to play the game. Their approach supports the idea that no single strategy can maintain high win rates against all other strategies.

Settlers of Catan is a similar board game that involves trades between players. In both, the action distribution is not uniform: certain action types (making trades) are more frequently valid than others. In Monopoly, a player is allowed to trade with other players at any point in the game. However, a player can only buy an unowned property (currently owned by the bank) when they land on the property square. \cite{dobre2017exploiting} use a model-based approach, Monte Carlo Tree Search (MCTS), for Settlers of Catan.  The authors in \cite{dobre2017exploiting} address the skewed action space by first sampling from a distribution over the types of legal actions followed by sampling individual actions from the chosen action type.

There is evidence emerging in other domains that hybrid DRL techniques reduce the computational complexity of the decision-making task and may provide a better alternative to a pure DRL approach. In \cite{wang2019lane}, authors combine DQN (for high-level lateral decision-making) with the rule-based constraints (to ensure safety) for autonomous driving to achieve a safe and efficient lane change behavior. The authors in \cite{wang2019lane} use the rules to ensure safety after the DRL has taken a decision. The rule-based method does not select an action instead it just returns whether it is safe to take the action returned by the DQN agent. \cite{chen2020combining} presents a framework for robots to pick up the objects in clutter by combining DRL and rule-based methods. They divide the task of picking up objects into two parts - pushing and grasping. They make use of the DDPG algorithm with a continuous output for pushing other objects out of the way and use a rule-based method to decide whether to grasp or not.  \cite{lee2020algorithm} propose an algorithm for the power-increase operation that uses an A3C agent for the continuous control module and a rule-based system for the discrete control components. 

In this work, we use DRL to solve decision-making in Monopoly. Like \cite{bailis, recent}, we represent Monopoly using an MDP, but unlike previous attempts, we do not simplify the game. To make the game as realistic as possible, we consider all possible actions (\Cref{table_action_type}), including trades. The inclusion of all actions makes the decision-making task more challenging since we need to deal with a high-dimensional action space. We also provide an improved state space representation and reward function when compared to the previous attempts. To handle the non-uniform action space, we propose a hybrid agent that combines a fixed-policy (or rule-based) approach for decisions involving rare actions with DRL for decisions involving remaining actions. Though the idea of hybrid DRL exists in literature, the separation of actions between DRL and fixed-policy is domain-specific. In \cite{chen2020combining, lee2020algorithm} this distinction is made on the basis that a subset of actions is continuous. In our case, the distinction is not obvious since all our actions are discrete. We are the first to propose the frequency of actions as a criterion for the selection of actions for rule-based and DRL respectively. In this work, we explicitly show that using a hybrid approach is better than using a standard one when all other variables are kept constant. In addition, we also show a comparison of using a policy gradient and value based method for our hybrid agent.
% \section{Game Representation}

\begin{table*}[ht]
\caption{Actions in Monopoly}
\label{table_action_type}
\centering
\begin{tabular*}{\textwidth}{@{\extracolsep{\fill}}llllr}
\toprule
%\begin{tabular}{|c|c|c|c|}
Action Type & Associated Properties & Game Phase &Action Parameters & Dimensions\\
\midrule
Make Trade Offer (Exchange) & All & Pre-roll, out-of-turn & \makecell[l]{To player, property offered, property requested,\\ cash offered, cash requested} & 2268 \Bstrut \\
\hline 
Make Trade Offer (Sell) & All & Pre-roll, out-of-turn & To player, property offered, cash requested & 252 \Tstrut \Bstrut \\
\hline 
Make Trade Offer (Buy) & All & Pre-roll, out-of-turn & To player, property requested, cash offered & 252 \Tstrut \Bstrut \\
\hline 
Improve Property & Color-group & Pre-roll, out-of-turn & Property, flag for house/hotel & 44 \Tstrut \Bstrut \\
\hline 
Sell House or Hotel & Color-group & Pre-roll, post-roll, out-of-turn & Property, flag for house/hotel & 44 \Tstrut \Bstrut \\
\hline 
Sell Property & All & Pre-roll, post-roll, out-of-turn & Property & 28 \Tstrut \Bstrut \\
\hline 
Mortgage Property & All & Pre-roll, post-roll, out-of-turn & Property & 28 \Tstrut \Bstrut \\
\hline 
Free Mortgage & All & Pre-roll, post-roll, out-of-turn & Property & 28 \Tstrut \Bstrut \\
\hline
Skip Turn & None & Pre-roll, post-roll, out-of-turn & None & 1 \Tstrut \Bstrut \\
\hline 
Conclude Actions & None & Pre-roll, post-roll, out-of-turn & None & 1 \Tstrut \Bstrut \\
\hline 
Use \textit{get out of jail} card & None & Pre-roll & None & 1 \Tstrut \Bstrut \\
\hline
Pay Jail Fine & None & Pre-roll & None & 1 \Tstrut \Bstrut \\
\hline 
Accept Trade Offer & None & Pre-roll, out-of-turn  & None & 1 \Tstrut \Bstrut \\
\hline 
Buy Property & All & Post-roll & Property & 1 \Tstrut \Bstrut \\
\bottomrule
\end{tabular*}
\end{table*}

% Below, we specify the action choice associated with each game phase:
% \begin{itemize}[leftmargin=*]
%     \item \textbf{Pre-roll Phase:} improve property, use get out of jail card, pay jail fine, skip turn, free mortgage, sell property, sell house or hotel, accept sell property offer, roll die, make trade offer, accept trade offer.
%     \item \textbf{Post-roll Phase:} buy property, sell property, sell house or hotel.
%     \item \textbf{Out-of-turn Phase:} free mortgage, sell property, sell house or hotel, accept sell property offer, make trade offer, accept trade offer, skip turn, improve property.
% \end{itemize} phase where it can take actions based on its new position after the dice roll. This phase is exclusive to the player who rolled the dice. 

\section{MDP Model for Monopoly}\label{sec_rep}
We design novel state and action space representations and utilize a combination of dense and sparse reward functions to model the full 4-player game of Monopoly as an MDP. 
\subsection{State Space}\label{sec_state}
We represent the state as a combination of player and property representation. For the player representation, we consider the current location, amount of cash with the player, a flag denoting if the player is currently in jail and another flag for whether the player has a \textit{get out of jail free} card. Since all other cards force a player to take an action immediately and are not part of the decision-making process, we do not consider them in the state space. For the property representation, we include the 28 property locations. These constitute 22 real-estate properties, four railroad properties, and two utility properties. The property representation consists of owner representation, a flag for a mortgaged property, a flag denoting whether the property is part of a monopoly, and the fraction of the number of houses and hotels built on the property to the total allowed number. We represent the owner as a 4-dimensional one-hot-encoded vector with one index for each player with all zeros indicating the bank.  In Monopoly, one can only build a house or a hotel on properties that belong to a color group. Thus for the non-real-estate properties, these values are always zero. We do not include the other locations from the board (\Cref{fig_board}) as they do not warrant a decision to be taken by the agent.  Overall, the state space representation is a 240-dimensional vector: 16 dimensions for the player representation and 224 dimensions for the property representation. 
\subsection{Action Space} \label{sec_action}
We consider all actions that require a decision to be made by the agent. We do not include compulsory actions, like paying tax, moving to a specific location because of a chance card, or paying rent when you land on a property owned by another player. An exhaustive list of actions considered can be found in \Cref{table_action_type}. 

We broadly classify the actions in Monopoly into three groups, those associated with all 28 properties, 22 color-group properties, or no properties. We represent all actions that are not associated with any properties as binary variables. Note that we only have one action that pertains to using a card: Use \emph{get out of jail} card. We do not consider any other card since all the other cards require the player to take the action mentioned on the card immediately. Only a \emph{get out of jail} card can be used at a later time, requiring the player to decide on when to use it. Since improvements (building a house or a hotel) in Monopoly are only allowed for properties belonging to a color group, we represent both \textit{improve property} and \textit{sell house or hotel} as a 44-dimensional vector where 22 dimensions indicate building a house with the other 22 indicating building a hotel on a given property. Actions that are associated with all properties, except for \textit{buy property} and \textit{make trade offer}, are represented using a 28-dimensional one-hot-encoded vector with one index for each property. A player is only allowed to buy an unowned property when they directly land on the property square. Hence, though the action is associated with all properties, the decision to buy or not can be represented using a binary variable.

Trades are possibly the most complex part of the game. A player is allowed to trade with other players anytime during the game. A trade offer has multiple parameters associated with it: it needs to specify the player to whom the trade is being offered. It may further include an offered property, a requested property, the amount of cash offered, and the amount of cash requested. We divide the trade offers into three sub-actions: \textit{sell property trade offers}, \textit{buy property trade offers}, and \textit{exchange property trade offers}. For the buy/sell trade offers, we discretize the cash into three parts: below market price (0.75 x purchase price), at market price (1 x purchase price) and, above market price (1.25 x purchase price). Since we have three other players, 28 properties, and three cash amounts, we represent these using a 252-dimensional (3x28x3) vector. To keep the dimensions in check for exchange trade offers, we use the market price for both assets. Thus, we only need to account for the properties and the player. We represent the exchange trade offers using a 2268-dimensional (3x28x27) vector. Altogether, the action space has 2922 dimensions. 

One thing to note here is that not all actions are valid all the time. Depending on the phase (\Cref{sec_simulator}) of the game, only a subset of possible actions is allowed (\Cref{table_action_type}).
\subsection{Reward Function}\label{sec_reward}
We use a combination of a dense and a sparse reward function (Eq. \eqref{eq:overall_reward}). In order to reward or penalize a player for the overall policy at the end of each game, we use a constant value of $\pm c$ for a win/loss respectively.
\begin{equation}
\label{eq:overall_reward}
    r = \begin{cases}
    \ \ \ \pm c\ \ \ \ \text{for a win/loss}
    \\ \ \ \ r_x \ \ \ \ \text{if the game is not over}
    \end{cases}
\end{equation}
where $r_x$ is the in-game reward for player $x$ and $c$ is a constant scalar. In order to choose the value of $c$, we ran an experiment with different values and chose the one that gives us the best results in terms of convergence time and the number of wins. More details are presented in \Cref{exp_results}.

During a single game, we use a reward function (Eq. \eqref{eq:reward}) defined as the ratio of the current players' net-worth (Eq. \eqref{eq:nw}) to the sum of the net worth of other active players. We update the net worth of each active player after they take any action.

\begin{equation}
\label{eq:nw}
nw_x=c_x + \sum_{a\in A_x} p_{a}
\end{equation}
where $nw_x$ is the net worth of player $x$, $c_x$ is the current cash with player $x$, $p_a$ is the price of asset $a$, and $A_x$ is the set of assets owned by player $x$. The value for $p_a$ is calculated as:
\begin{equation}
\label{eq:pa}
p_a = (b_p - m_v) * b + n_h * p_h + n_H + p_H
\end{equation}
where $b_p$ is the base price (as shown in \Cref{fig_board}) of an asset, $m_v$ is the mortgage value, $b$ denotes a bonus constant, $n_h$ and $p_h$ denote the number of houses and the price of each house on a given property and, $n_H$ and $p_H$ denote the number of hotels and price of each hotel built on the property. If an asset is mortgaged, $m_v$ equals the mortgaged owed on the property. If the asset is not mortgaged, then this value is zero. Since, in Monopoly, it is often better to have property than have cash of the same value, we give a bonus constant to encourage the agent to buy more properties. $b = 1.5$ if the asset is not part of a Monopoly and $b = 2$ if the asset is part of a Monopoly.
For the in-game reward we have:
\begin{equation}
\label{eq:reward}
r_x = \frac{nw_x}{{\sum_{y\in X_i \setminus x}}nw_y}
\end{equation}
where $r_x$ is the in-game reward for player $x$ and $X$ is the set of all active players. This reward value is bounded between [0,1] and helps distinguish the relative value of each state-action pair within a game. 
\section{Approach}\label{sec_approach}
We approach Monopoly from a single agent perspective and treat the other players as part of the environment. We adopt a DRL approach to tackle the decision-making task. As we saw in the previous section (\Cref{sec_action}), there are multiple actions an agent can take at any given stage of the game resulting in a complex learning problem. We propose two learning-based approaches, a policy gradient method (PPO) and a value based method (DDQN). In both cases, we train a standard agent that uses a model-free DRL paradigm for all decisions and a hybrid agent that uses DRL for a subset of actions in conjunction with a fixed-policy for remaining actions.
\subsection{Standard DRL Agents} \label{sec_standard}
\subsubsection{Actor-Critic PPO Agent}
We implement the actor-critic PPO algorithm with a clipped surrogate objective function \cite{schulman2017proximal}. To estimate the advantage, we make use of the truncated version of generalized advantage estimation used in \cite{mnih2016asynchronous}. The actor-critic PPO implementation makes use of two independent networks - the actor network and the critic network. PPO is an on-policy algorithm in which the agent follows a policy dictated by the actor network for a fixed number of time-steps (much less than the episode length). The actor network parameters are initialized randomly, and the agent \textit{explores} by sampling actions according to the latest version of its stochastic policy. As the training proceeds, the policy typically becomes less random since the update rule encourages it to \textit{exploit} rewards. The critic network is responsible for calculating the value of a given state which is used in the advantage estimation when optimizing the objective function. More details on the approach are presented in Appendix \ref{appendix_ppo_approach}.

\subsubsection{DDQN Agent}
A common issue with using vanilla DQN is that it tends to overestimate the expected return. Double Q-learning \cite{hasselt2010double} overcomes this problem by making use of a double estimator. To avoid over-estimation of the Q-values, we implement the DDQN \cite{van2016deep} algorithm to train our agent. Similar to the standard DQN approach, DDQN makes use of an experience replay \cite{lin1992self} and a target network. Similar to \cite{mnih2015human}, we make use of the $\epsilon$-greedy exploration policy to select actions. Initially, the agent \textit{explores} the environment by randomly sampling from allowed actions. As the learning proceeds and the agent learns which actions are more successful, its exploration rate decreases in favor of more \textit{exploitation} of what it has learned. We mask the output of the network to only the allowed actions to speed up the training process. The action masking ensures that the learning agent selects a valid action at any given time. More details on the approach are presented in Appendix \ref{appendix_ddqn_approach}.

\subsection{Hybrid Agents}\label{sec_hybrid}
Standard DRL techniques have a high sample complexity. DRL requires each state-action pair to be visited infinitely often, the main reason we use $\epsilon$-greedy in the DDQN agent. If some states are rare, we do not want to force the agent to explore them - especially if the related decisions are straightforward and we have an idea of what actions might be good in the given state. When playing Monopoly, a player can only buy an unowned property (property still owned by the bank) when they exactly land on the property square. During our simulations, we observed that the \textit{buy property} action is seldom allowed. Similarly, \textit{accept trade offer} is only valid when there is an outstanding trade offer from another player. The resulting rare-occurring action-state pairs further increase the sample and computational complexity of the learning task. We hypothesize that by using a rule-based approach for the rare occurring but simple decisions and a learning-based approach for the more frequent but complex decisions, we can improve the overall performance.

We design hybrid agents that integrate the DRL approaches presented earlier (\Cref{sec_standard}) with a fixed-policy approach. We use a fixed-policy to make \textit{buy property} and \textit{accept trade offer} decisions. For all other decisions, we use DRL. During training, if there is an outstanding trade offer, the execution flow shifts from the learning-based agent to a fixed-policy agent to decide whether to accept the trade offer or not. The agent accepts an offer if the trade increases the number of monopolies. If the number of monopolies remains unchanged, the agent only accepts if the net worth of the offer is positive. The net worth of an offer is calculated using:
\begin{equation}\label{eq_trade}
nw_o = (p_o + c_o) - (p_r + c_r)
\end{equation} where $nw_o$ denotes the net worth of the trade offer, $p_o$ is the price of the property offered, $c_o$ is the amount of cash offered, $p_r$ is the price of the property requested, and $c_r$ is the amount of cash requested. 

Similarly, whenever the agent lands on a property owned by the bank, the fixed-policy agent decides whether or not to buy the property. The agent buys the property if it results in a monopoly as long as it can afford it. For all other properties, if the agent has \$200 more than the property price, it decides to buy. Our experiments show that in both cases, the hybrid agent converges faster and significantly outperforms the standard DRL agent when playing against other fixed-policy agents.

\section{Experimental Setup}

\subsection{Monopoly Simulator}\label{sec_simulator}
We develop an open-sourced, complete simulator for a four-player game of Monopoly using Python, available on GitHub\footnote{\url{https://github.com/mayankkejriwal/GNOME-p3}}. The simulator implements the conventional Monopoly board with 40 locations shown in \Cref{fig_board} and enforces rules similar to the US version of the game\footnote{\url{https://www.hasbro.com/common/instruct/monins.pdf}}, barring some modifications. We do not consider the game rules associated with rolling doubles (for example, a double can get a player out of jail). We treat them as any other dice roll. Trading is an integral part of Monopoly. Players can use trades to exchange properties with or without cash with one or more players. We enforce the following rules for trading:
\begin{itemize}
    \item Players can trade only unimproved (no houses or hotels) and unmortgaged properties.
    \item Players can make trade offers simultaneously to multiple players.  The player who receives a trade offer is free to either accept or reject it. The trade transaction gets processed only when a player accepts an offer. Once a trade transaction is processed, we terminate all other simultaneous trade offers for the same property.
    \item A player can have only one outstanding trade offer at a time. A player needs to accept or reject a pending offer before another player can make a different trade offer. 
\end{itemize}

In the conventional setting, players can take certain actions like mortgaging or improving their property even when it is not their turn to roll dice. If multiple players take simultaneous actions, the game could become unstable. To avoid this and to be able to keep track of all the dynamic changes involved in the game, we divide the gameplay into three phases:  
\begin{itemize}
\item Pre-roll: The player whose turn it is to roll the dice is allowed to take certain actions before the dice roll in this phase. To end the phase, the player needs to \textit{conclude actions}.
\item Out-of-turn: Once the pre-roll phase ends for a player,  the other players can make some decisions before this player rolls the dice. Every player is allowed to take actions in a round-robin manner in this phase until all players decide to \textit{skip turn} or a predefined number of out of turn rounds are complete.
\item Post-roll: Once the player rolls dice, their position is updated based on the sum of the number on the dice. This player then enters the post-roll phase. If the player lands on a property that is owned by the bank, they need to decide whether or not to buy during this phase.
\end{itemize}
\Cref{table_action_type} shows the game phases associated with each action. If a player has a negative cash balance at the end of their post-roll phase, they get a chance to amend it. If they are unsuccessful in restoring the cash balance, bankruptcy procedure begins following which the player loses the game.

\subsection{Baseline Agents}\label{sec_baseline}
We develop baseline agents that, in addition to buying or selling properties, can make trades. We base the policies of these agents on successful tournament-level strategies adopted by human players. Several informal sources on the Web have documented these strategies though they do not always agree\footnote{Two resources include \url{http://www.amnesta.net/monopoly/} and \url{https://www.vice.com/en/article/mgbzaq/10-essential-tips-from-a-monopoly-world-champion}.}. A complete academic study on which strategies yield the highest probabilities of winning has been lacking. Perhaps the complex rules of the game have made it difficult to formalize analytically.

We develop three fixed-policy agents: FP-A, FP-B, and FP-C. All three agents can make one-way (buy/sell) or two-way (exchange) trades with or without cash involvement. They are also capable of rolling out trade offers simultaneously to multiple players. By doing so, the agent increases the probability of a successful trade, so it can acquire properties that lead to monopolies of a specific color group more easily. The fixed-policy agents try to offer properties that hold a low value (for example, a solitary property) to the agent itself but may be of value (gives the other player a monopoly) to the other player and vice versa when making trade requests. To yield a higher cash balance, the agents seek to improve their monopolized properties by building houses and hotels.

All three agents place the highest priority in acquiring a monopoly but differ on the priority they base on each property. FP-A gives equal priority to all the properties, FP-B and FP-C give a high priority to the four railroad properties. Additionally, FP-B places a high priority on the high rent locations: \textit{Park Place} and \textit{Boardwalk} and assigns a low priority to utility locations. On the other hand, FP-C places high priority on properties in the orange color group (\textit{St. James Place}, \textit{Tennessee Avenue}, \textit{New York Avenue}) or in the sky-blue color group (\textit{Oriental Avenue}, \textit{Vermont Avenue}, \textit{Connecticut Avenue}). An agent tries to buy or trade properties of interest more aggressively, sometimes at the risk of having a low cash balance. It may also end up selling a lower priority property to generate cash for a property of interest.

\section{Experiments and Results}\label{exp_results}
\begin{figure}[!t]
\centering
\includegraphics[width=0.45\textwidth]{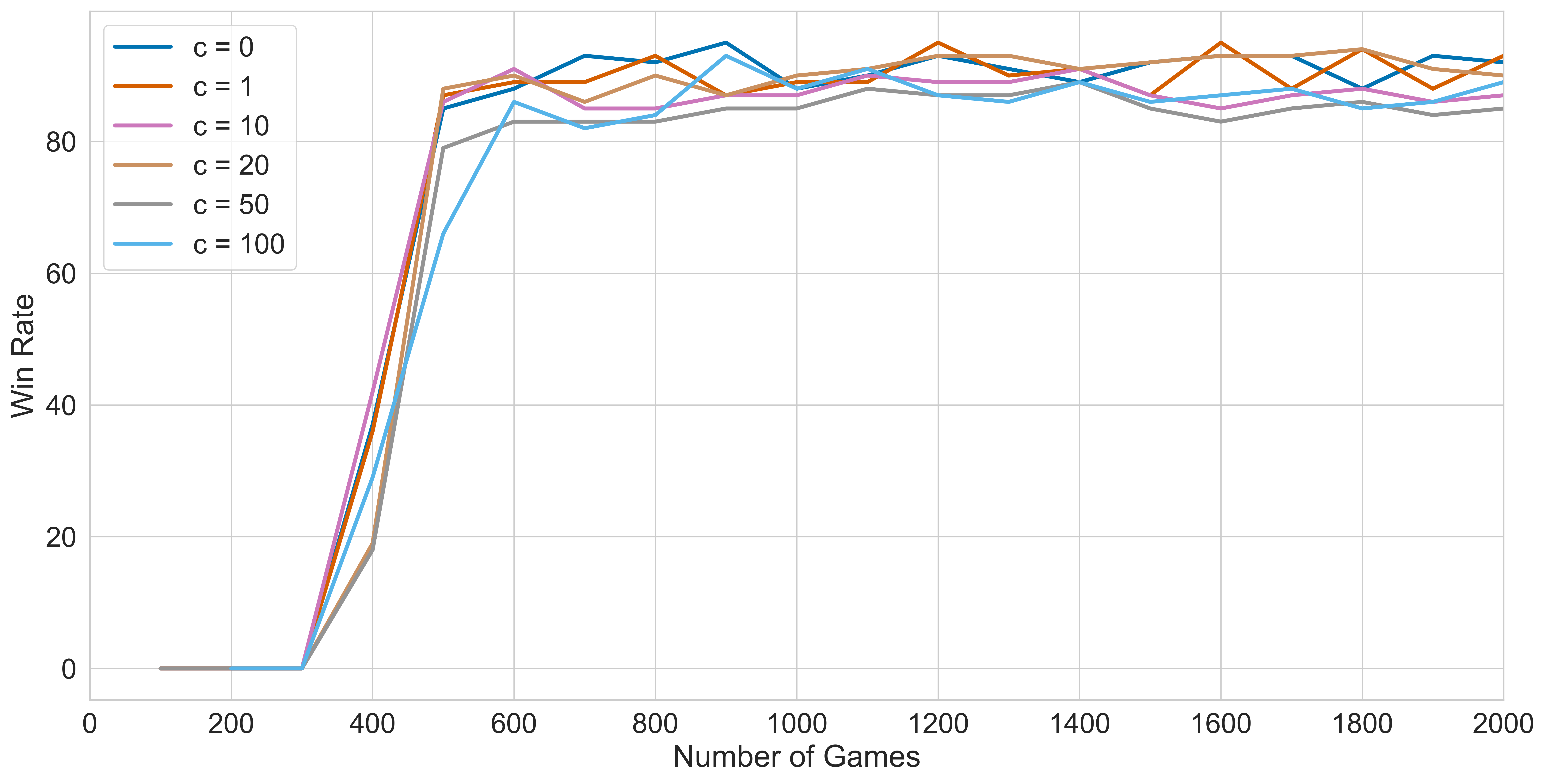}
\caption{Comparison of win rate (number of wins every 100 games) for different values of $c$ in the reward function \Cref{eq:overall_reward} for hybrid PPO agent.}
\label{fig_ppo_c}
\end{figure}

\begin{figure}[!t]
\centering
\includegraphics[width=0.45\textwidth]{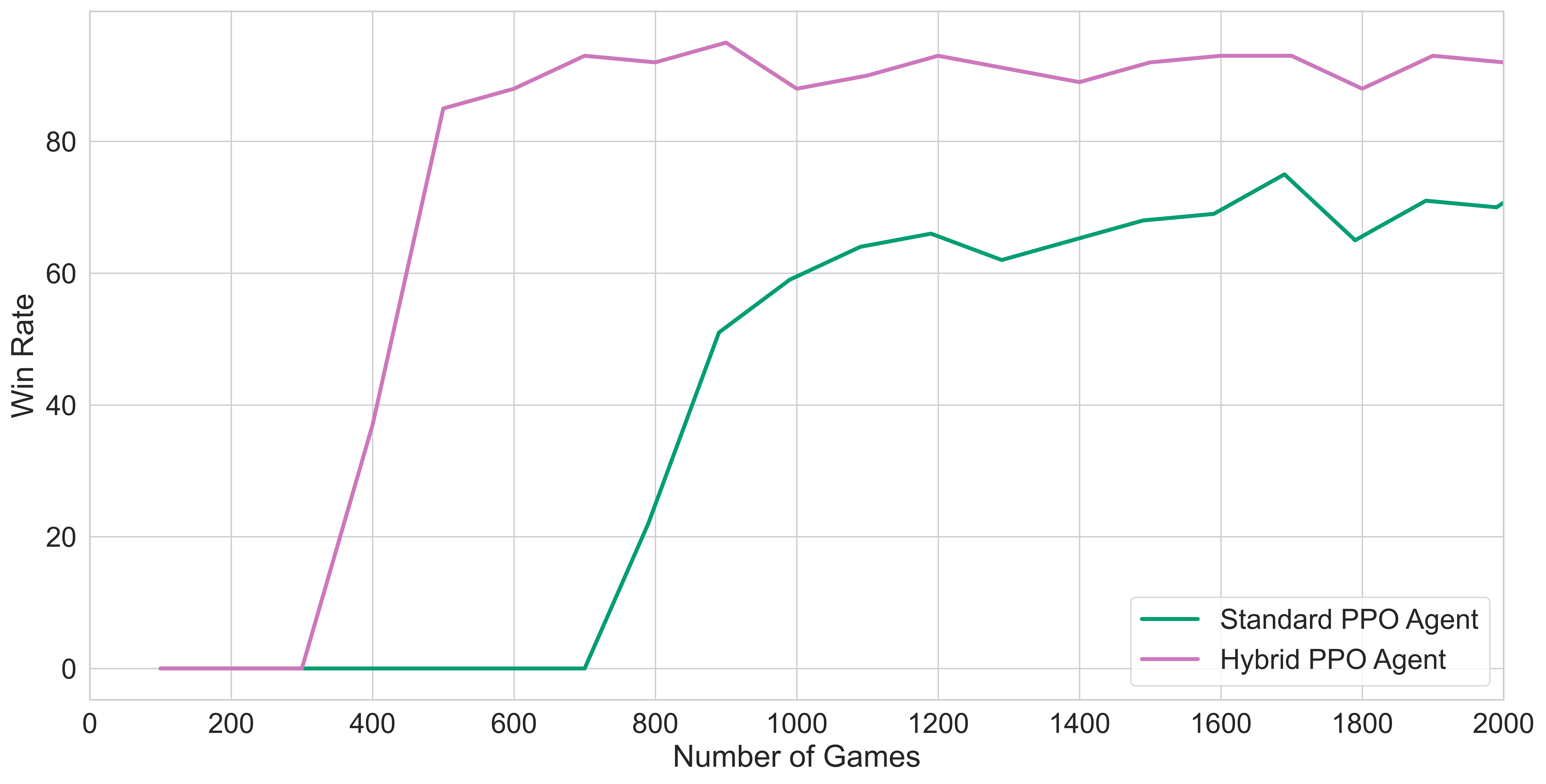}
\caption{Comparison of win rate (number of wins every 100 games) for standard PPO and hybrid PPO agent during training. The hybrid agent converges sooner and to a better policy as compared to the standard agent.}
\label{fig_wins_ppo}
\end{figure}

\color{black}
\subsection{Training of PPO Agents:}
We train both the standard PPO agent and the hybrid PPO agent using the actor critic PPO algorithm. We use the same architecture and parameters for both agents in order to draw a fair comparison. We use the state, action, and reward functions presented in \Cref{sec_rep}. In the case of the hybrid agent, however, we permanently mask the actions that use a fixed-policy. During training, the agents play against the three fixed-policy agents. We train the learning agents for 2000 games each and randomize the turn order to remove any advantage one may get due to the player's position. Details on the network architecture and parameters are provided in Appendix \ref{net_param_ppo}.\\

% \subsubsection*{\textcolor{red}{Network Architecture and Parameters}} We use a fully connected feed-forward network to approximate $Q(s_t,a_t)$  for the policy network. The input to the network is the current state of the environment, $s_t$, represented as a 240-dimensional vector (\Cref{sec_state}). We make use of 2 hidden layers, that consist 1024 and 512 neurons respectively, each with a rectified linear unit (ReLU) as the activation function:

% \begin{equation}
%     f(x) = \begin{cases}
%     x\ \ \ \ \text{for}\ x\geq 0
%     \\ 0 \ \ \ \ \text{otherwise}
%     \end{cases}
% \end{equation}

% The output layer has a dimension of $2922$, where each element represents the Q-value for each of the actions the agent can take. As discussed earlier, not all actions are valid at all times. We mask the output of the final layer to only the allowed actions. For training the network, we employ the Adam optimizer \cite{kingma2014adam} and use mean-square error as the loss function. We initialize the target network with the same architecture and parameters as the policy network. We update the parameter values of the target network to that of the policy network every 500 episodes and keep them constant otherwise. After tuning our network, we achieved the best results using the following parameters: $\gamma = 0.9999$ , learning rate $\alpha = 10^{-5}$ , batch size $ = 128$, and a memory size $ = 10^4$.\\\\
\noindent \textbf{\noindent Experiment 1 - Constant value for reward}\\
In order to see the effect of the constant value $c$ on the performance of the agent, we train the hybrid PPO agent with different values of $c \in \{0,1, 10, 20, 50, 100\}$. The win rate (wins per 100 games) is given in \Cref{fig_ppo_c}. Since PPO is an on-policy algorithm and relies more on the in-game reward, we see that the value of $c$ (received during win/loss) does not have a major impact on the performance of the agent. We keep $c=0$ for the PPO agents for all the following experiments.\\\\
\noindent \textbf{\noindent Experiment 2 - Standard PPO vs Hybrid PPO}\\
We compare the performance of the standard PPO agent to that of the hybrid PPO agent during training. The win rate (wins per 100 games) for each agent during training is shown in \Cref{fig_wins}. As mentioned earlier, the difference between the two settings is that for the standard PPO agent, all the decisions are taken by the learning agent, whereas for the hybrid PPO agent, a subset of decisions is taken by following a fixed-policy approach. We see from the graph that the hybrid PPO agent converges sooner and to a better policy when compared to the standard PPO agent. 
\color{black}
\subsection{Training of DDQN Agents} \label{sec_training}
Previous attempts at Monopoly \cite{bailis, recent} use Q-value based methods to solve the MDP task. In order to compare our approach to the previous approaches\footnote{We omit the comparison of action space representations since the agent does not win a single game against the fixed-policy agents when using a simpler action space presented in \cite{bailis, recent}.} and to see how a value based algorithm behaves, we train both the standard agent and the hybrid agent using the DDQN algorithm. For the two agents to be comparable, we use the same architecture and parameters for both. We use the state, action, and reward functions presented in \Cref{sec_rep}. Like in the PPO case, we permanently mask the actions that use a fixed-policy for the hybrid DDQN agent. Details on the network architecture and parameters are provided in Appendix \ref{net_param_ddqn}.\\

\begin{figure}[!t]
\centering
\includegraphics[width=0.45\textwidth]{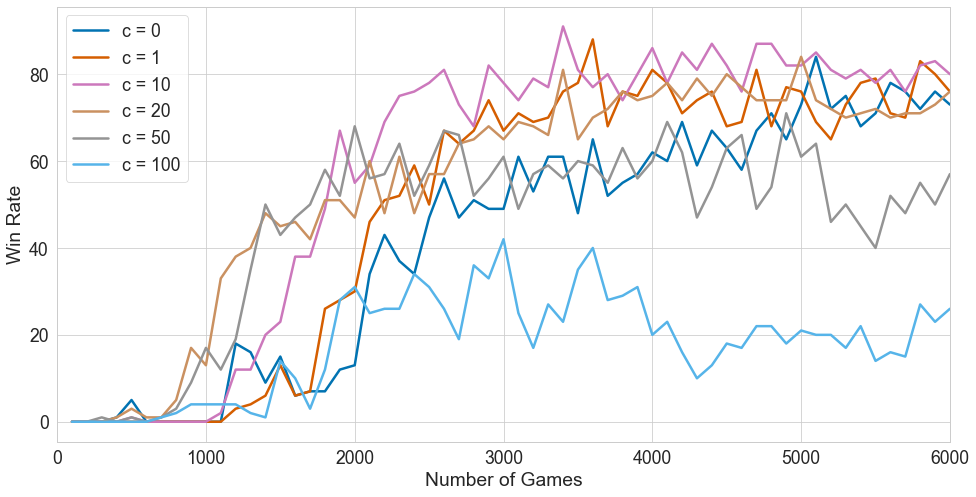}
\caption{Comparison of win rate (number of wins every 100 games) for different values of $c$ in the reward function \Cref{eq:overall_reward} for hybrid DDQN agent.}
\label{fig_c}
\end{figure}

\color{black}
\noindent \textbf{Experiment 1 - Constant value for reward}\\
In order to see the effect of the value of the constant $c$ for the reward function (\Cref{eq:overall_reward}), we run an experiment with different values of $c \in \{0,1, 10, 20, 50, 100\}$ when training the hybrid agent. The hybrid agent is trained for 6000 games against the three fixed-policy agents for each value of $c$. We present the results for the win rate (wins per 100 games) in  \Cref{fig_c}. Since DDQN is an off-policy algorithm and uses a larger experience buffer, we see that, unlike the PPO agent, the value of $c$ affects the performance of the DDQN agent. We get the fastest convergence and best win rates on average for $c=10$. For all the following experiments we fix the value of $c$ at 10.\\

\color{black}

\begin{figure}[!t]
\centering
\includegraphics[width=0.45\textwidth]{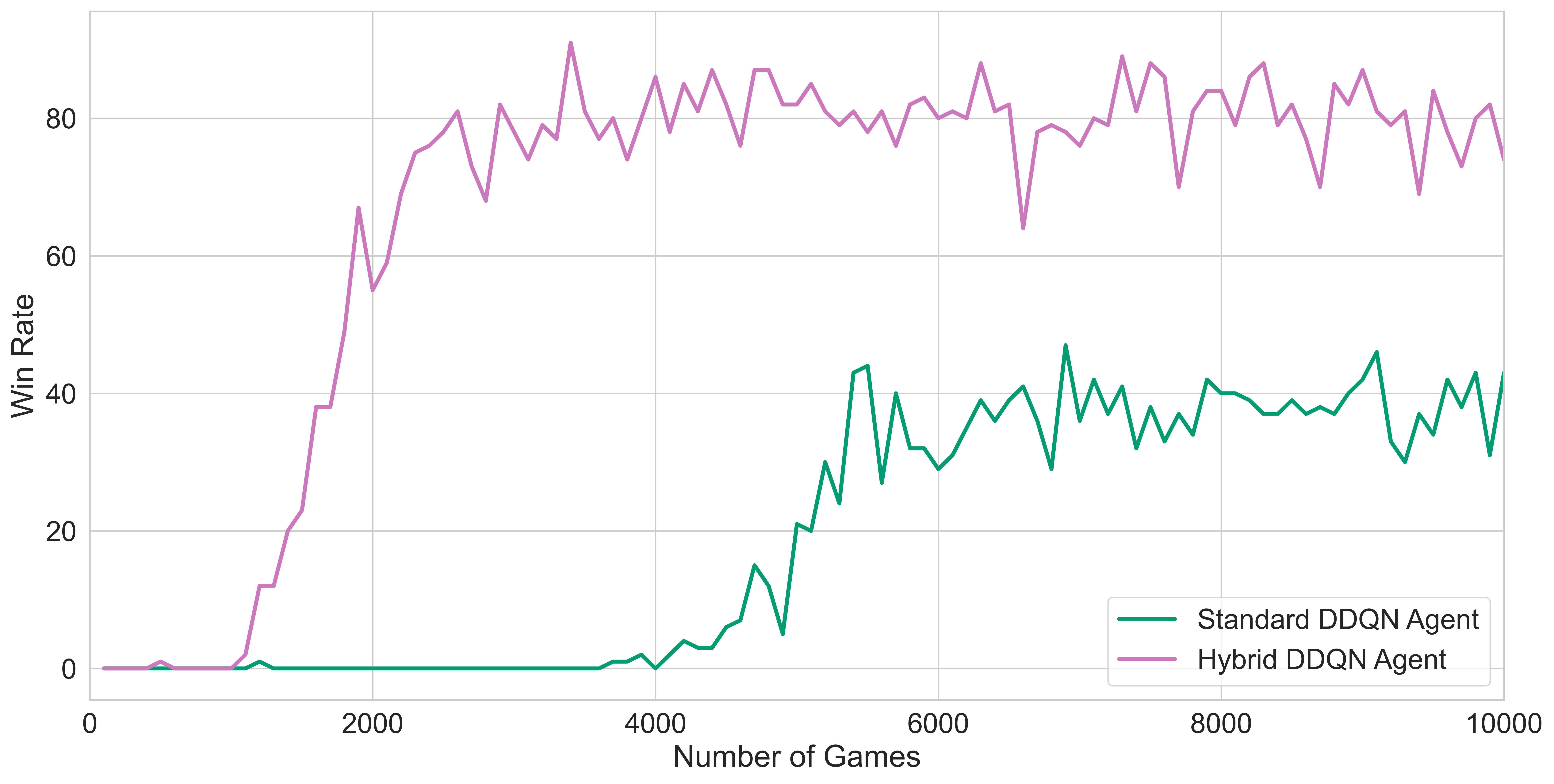}
\caption{Comparison of win rate (number of wins every 100 games) for standard DDQN and hybrid DDQN agent during training. The hybrid agent converges sooner and to a better policy as compared to the standard DRL agent.}
\label{fig_wins}
\end{figure}

\begin{figure}[!t]
\centering
\includegraphics[width=0.45\textwidth]{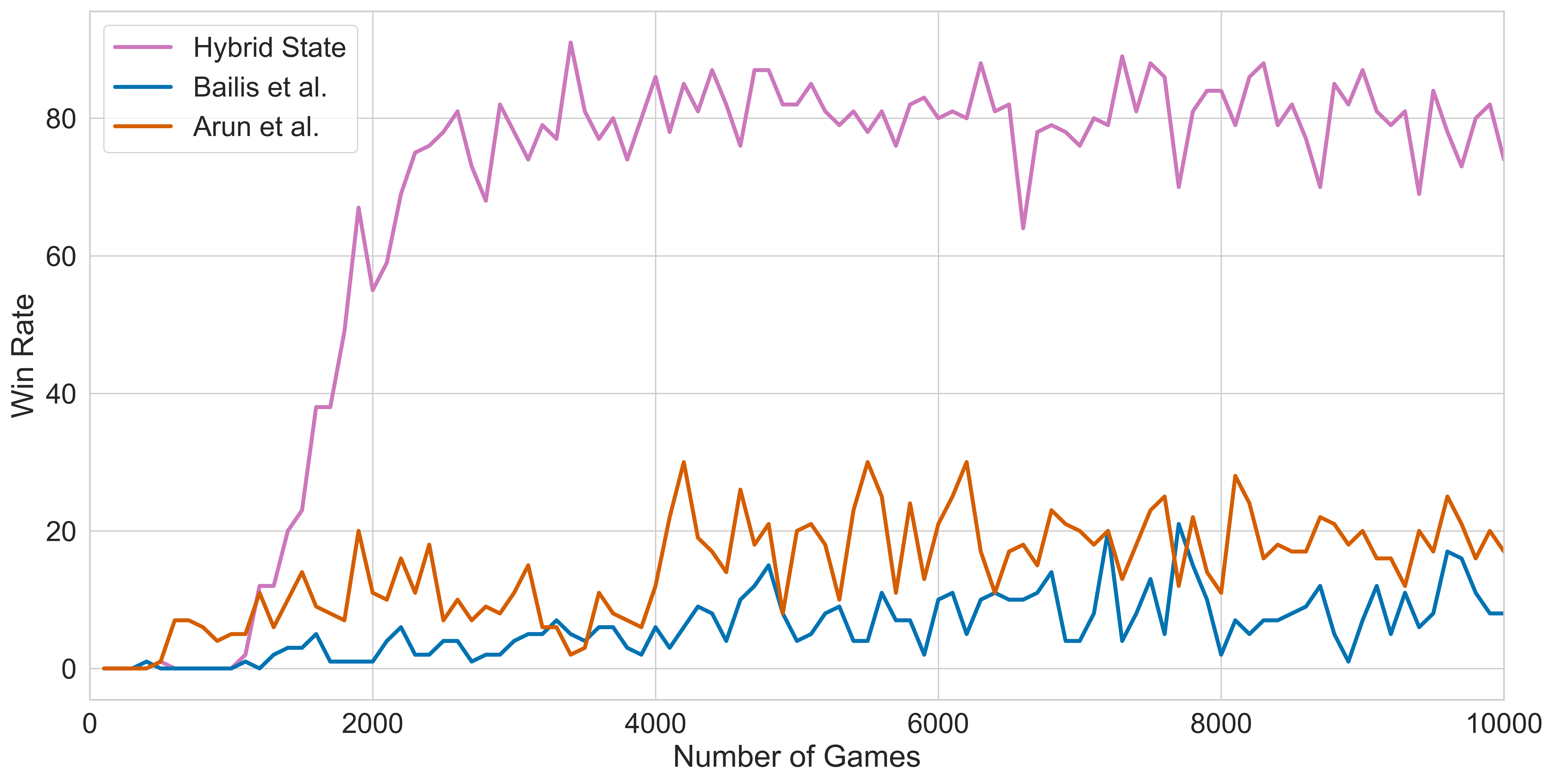}
\caption{Comparison of win rates of the hybrid DDQN agent during training using our proposed state space representation (\Cref{sec_state}) to that previously given by Bailis et al. \cite{bailis} and Arun et al. \cite{recent}.}
\label{fig_statecomp}
\end{figure}

\begin{figure}[!t]
\centering
\includegraphics[width=0.45\textwidth]{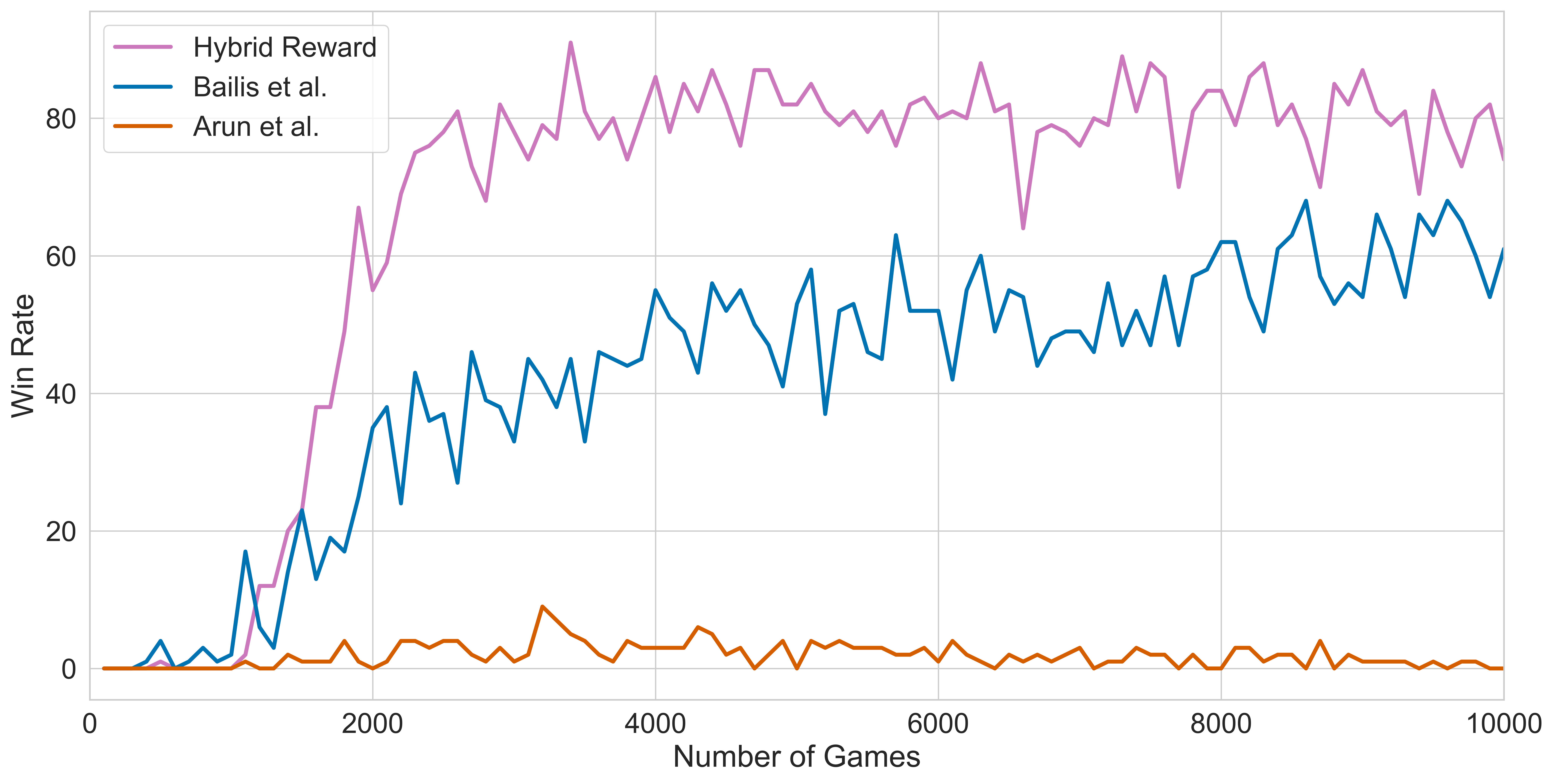}
\caption{Comparison of win rates of the hybrid DDQN agent during training using our proposed reward function (\Cref{sec_reward}) to that previously given by Bailis et al. \cite{bailis} and Arun et al. \cite{recent}.}
\label{fig_rewardcomp}
\end{figure}

\noindent \textbf{Experiment 2 - Standard DDQN vs Hybrid DDQN}\\
In order to compare the standard DRL agent to the hybrid agent, we run an experiment to show how the two agents behave during training and show a comparison of the win rate while training. During training, the agents play against the three fixed-policy agents. We train the learning agents for 10000 games each and use an exponential decay function for the exploration rate. We randomize the turn order during training (and testing) to remove any advantage one may get due to the player's position. The win rate (wins per 100 games) for each agent during training is shown in \Cref{fig_wins}.\\\\ 
\color{black}
\noindent \textbf{Experiment 3 - Comparison of state space representations}\\
We perform an experiment to show the comparison of using different state space definitions for the DDQN agent. We compare our state space definition \Cref{sec_state} to the state space representation proposed by Bailis et al. \cite{bailis} and Arun et al. \cite{recent}. \Cref{fig_statecomp} shows a comparison of win rates of the hybrid agent during training using the three different state representations. Please note, we use our action space representation (\Cref{sec_action})  and reward function (\Cref{sec_reward}) for all three training runs and only change the state space representation.
\\\\
\noindent \textbf{ Experiment 4 - Comparison of reward functions}\\
We run an experiment to show how the choice of the reward function affects the performance of the learning agent. We compare our reward function \Cref{sec_reward} to the reward functions proposed by Bailis et al. \cite{bailis} and Arun et al. \cite{recent}. \Cref{fig_rewardcomp} shows a comparison of the win rates of the hybrid agent during training using the three different reward functions. For these training runs, we use our state and action space representations and only change the reward functions. 
\color{black}

\subsection{Evaluation of Learning Agents}\label{sec_results}

We run multiple experiments to evaluate the different learning agents. In the case of the standard PPO agent, all decisions are made using the pre-trained actor network. The actor network makes a subset of the decisions for the hybrid PPO agent and follows a fixed-policy for the remaining actions. We use the pre-trained policy network to take all the decisions in the case of the standard DDQN agent and a subset of decisions in the case of the hybrid DDQN agent. We set the exploration rate to zero for both the standard and the hybrid DDQN agents. For each experiment, we use different combinations of agents to play the four-player Monopoly game. Each experiment is run for five iterations of 2000 games each. The order of play is randomized for each game.\\
\color{black}

\noindent \textbf{\noindent Experiment 1 - PPO vs Fixed-Policy}\\
For the first experiment, we evaluate the performance of the two PPO agents when playing against the three fixed-policy agents. We provide the results for the standard PPO agent and the hybrid PPO agent in \Cref{table_rl_results} and \Cref{table_hybrid_results} respectively. We observe that the standard PPO agent has a win rate of \textbf{69.95\%} while that of the hybrid PPO agent is \textbf{91.65\%}. The result is in line with what we saw during training. The hybrid PPO agent outperforms the standard PPO agent by more than \textbf{20\%} when playing against fixed-policy agents.\\
\color{black}
\begin{table}[!hbpt]
% \renewcommand{\arraystretch}{1.3}
% if using array.sty, it might be a good idea to tweak the value of
% \extrarowheight as needed to properly center the text within the cells
\color{black}
\caption{Results for standard PPO agent against fixed-policy agents over five runs of 2000 games each}
\label{table_ppo_rl_results}
\centering
\begin{tabular}{ccccc}
\toprule 
Run & FP-A & FP-B & FP-C & Std PPO\\
\midrule
1 & 172 & 177 & 278 & 1373 \Bstrut \\
2 & 157 & 183 & 279 & 1381 \Bstrut \\
3 & 125 & 174 & 275 & 1426 \Bstrut \\
4 & 161 & 158 & 270 & 1411 \Bstrut \\
5 & 154 & 187 & 255 & 1404 \Bstrut \\
\midrule
Win Rate & 7.69\% & 8.79\% & 13.57\% & \textbf{69.95\%} \\
\bottomrule
\end{tabular}
\end{table}
\begin{table}[!hbpt]\color{black}
\caption{Results for hybrid PPO agent against fixed-policy agents over five runs of 2000 games each}
\label{table_ppo_hybrid_results}
\centering
\begin{tabular}{ccccc}
\toprule 
Run & FP-A & FP-B & FP-C & Hybrid PPO\\
\midrule
1 & 35 & 53 & 68 & 1844 \Bstrut \\
2 & 42 & 47 & 68 & 1843 \Bstrut \\
3 & 48 & 51 & 69 & 1832 \Bstrut \\
4 & 32 & 68 & 80 & 1820 \Bstrut \\
5 & 50 & 69 & 55 & 1826 \Bstrut \\
\midrule
Win Rate & 2.07\% & 2.88\% & 3.40\% & \textbf{91.65\%} \\
\bottomrule
\end{tabular}
\end{table}

\newpage
\noindent \textbf{\noindent Experiment 2 - DDQN vs Fixed-Policy}\\
We perform a similar evaluation for the DDQN agents as we do for PPO agents in the first experiment. We run five iterations of 2000 games each against the three fixed-policy agents for both learning agents. The standard DRL agent achieves a win rate of \textbf{47.41\%} as shown in Table \ref{table_rl_results}. The hybrid agent significantly outperforms the standard agent and achieves a win rate of \textbf{76.91\%} as shown in Table \ref{table_hybrid_results}.\\

\begin{table}[!hbpt]
% \renewcommand{\arraystretch}{1.3}
% if using array.sty, it might be a good idea to tweak the value of
% \extrarowheight as needed to properly center the text within the cells
\caption{Results for standard DDQN agent against fixed-policy agents over five runs of 2000 games each}
\label{table_rl_results}
\centering
\begin{tabular}{ccccc}
\toprule 
Run & FP-A & FP-B & FP-C & Std DDQN\\
\midrule
1 & 307 & 307 & 424 & 962 \Bstrut \\
2 & 307 & 298 & 455 & 940 \Bstrut \\
3 & 286 & 332 & 430 & 952 \Bstrut \\
4 & 267 & 333 & 451 & 949 \Bstrut \\
5 & 329 & 331 & 402 & 938 \Bstrut \\
\midrule
Win Rate & 14.96\% & 16.01\% & 21.62\% & \textbf{47.41\%} \\
\bottomrule
\end{tabular}
\end{table}
\begin{table}[!hbpt]
\caption{Results for hybrid DDQN agent against fixed-policy agents over five runs of 2000 games each}
\label{table_hybrid_results}
\centering
\begin{tabular}{ccccc}
\toprule 
Run & FP-A & FP-B & FP-C & Hybrid DDQN\\
\midrule
1 & 148 & 169 & 147 & 1536 \Bstrut \\
2 & 143 & 172 & 142 & 1543 \Bstrut \\
3 & 147 & 168 & 134 & 1551 \Bstrut \\
4 & 161 & 176 & 124 & 1539 \Bstrut \\
5 & 147 & 186 & 145 & 1522 \Bstrut \\
\midrule
Win Rate & 7.46\% & 8.71\% & 6.92\% & \textbf{76.91\%} \\
\bottomrule
\end{tabular}
\end{table}

\color{black}
\noindent \textbf{\noindent Experiment 3 - Hybrid vs Standard vs Fixed-Policy}\\
To test how the standard and the hybrid agents perform against each other, we evaluate a setting where the standard and hybrid agents play against each other as well against two fixed-policy agents. We present the results from one of those settings in \Cref{table_ppo_v45} for the PPO agents and in \Cref{table_ppo_v45} for the DDQN agents. In both cases, we observe that the hybrid agents have a higher win rate (\textbf{88.72\%} and \textbf{56.86\%}) than the standard agents (\textbf{5.97\%} and \textbf{13.87\%}). We present results for additional settings in Appendix \ref{appendix_add_eval_results}. \\

\begin{table}[!hbpt]\color{black}
\caption{Results for hybrid and standard PPO agents against fixed-policy agents over five runs of 2000 games each}
\label{table_ppo_v45}
\centering
\begin{tabular}{ccccc}
\toprule 
Run & FP-B & FP-C & Std PPO & Hybrid PPO\\
\midrule
1 & 57 & 40 & 108 & 1795 \Bstrut \\
2 & 53 & 51 & 135 & 1761 \Bstrut \\
3 & 54 & 60 & 107 & 1779 \Bstrut \\
4 & 46 & 52 & 113 & 1789 \Bstrut \\
5 & 68 & 50 & 134 & 1748 \Bstrut \\
\midrule
Win Rate & 2.78\% & 2.53\% & 5.97\% & \textbf{88.72\%} \\
\bottomrule
\end{tabular}
\end{table}
\begin{table}[!hbpt]
\color{black}\caption{Results for hybrid and standard DDQN agents against fixed-policy agents over five runs of 2000 games each}
\label{table_v45}
\centering
\begin{tabular}{ccccc}
\toprule 
Run & FP-B & FP-C & Std DDQN & Hybrid DDQN\\
\midrule
1 & 259 & 299 & 272 & 1170 \Bstrut \\
2 & 326 & 293 & 268 & 1113 \Bstrut \\
3 & 300 & 299 & 261 & 1140 \Bstrut \\
4 & 287 & 269 & 293 & 1151 \Bstrut \\
5 & 287 & 308 & 293 & 1112 \Bstrut \\
\midrule
Win Rate & 14.59\% & 14.68\% & 13.87\% & \textbf{56.86\%} \\
\bottomrule
\end{tabular}
\end{table}

\noindent \textbf{\noindent Experiment 4 - PPO vs DDQN}\\
For our final evaluation, we evaluate how the four learning agents perform when playing against each other. We see from \Cref{table_ppo_v_ddqn_results} that the standard agents barely win any games. The hybrid PPO agent performs the best with a win rate of \textbf{69.06\%} and the hybrid DDQN comes in second with a win rate of \textbf{28.56\%}.
\color{black}
\begin{table}[!hbpt]
\color{black}\caption{Results for PPO agents against DDQN agents over five runs of 2000 games each}
\label{table_ppo_v_ddqn_results}
\centering
\begin{tabular}{ccccc}
\toprule 
Run & Std DDQN & Hybrid DDQN & Std PPO & Hybrid PPO\\
\midrule
1 & 18 & 566 & 27 & 1389 \Bstrut \\
2 & 15 & 594 & 24 & 1367 \Bstrut \\
3 & 23 & 559 & 33 & 1385 \Bstrut \\
4 & 17 & 592 & 32 & 1359 \Bstrut \\
5 & 17 & 545 & 32 & 1406 \Bstrut \\
\midrule
Win Rate & 0.9\% & 28.56\% & 1.48\% & \textbf{69.06\%} \\
\bottomrule
\end{tabular}
\end{table}
\subsection{Discussion}
From the results, we see that the learning agents outperform the fixed-policy agents by some margin. Although only two action choices separate the standard and the hybrid agents, we observe that the hybrid agents significantly outperform the standard agents. Evidently, instead of letting the agent explore the rare state-action pair it may be better suited if these are replaced by rule-based logic, especially if we know what actions might be good in the given state. %Thus, for a complex decision-making task like Monopoly, it may be best to use a hybrid approach if certain decisions occur less frequently than others: DRL for more frequent but complex decisions and a fixed policy for the less frequent but straightforward ones. 
Additionally, we see from \Cref{fig_wins_ppo} and \Cref{fig_wins}, the hybrid agents converge sooner and to a better policy than the standard DRL agents. From \Cref{fig_statecomp} and \Cref{fig_rewardcomp} we see that our state representation and design of reward function improves the performance of the DDQN learning agent when compared to previous attempts. Though we see a similar pattern for both value based and policy gradient based methods, we note that the hybrid PPO agent outperforms the hybrid DDQN agent.
\section{Conclusion}
We present the first attempt at modeling the full version of Monopoly as an MDP. Using novel state and action space representations and an improved reward function, we show that our DRL agent learns to win against different fixed-policy agents. The non-uniform action distribution in Monopoly makes the decision-making task more complex. To deal with the skewed action distribution we propose a hybrid DRL approach. The hybrid agents use DRL for the more frequent but complex decisions combined with a fixed-policy for the infrequent but simple decisions. Experimental results show that the hybrid agents significantly outperform standard DRL agents for both policy based (PPO) and value based methods (DDQN). We evaluate the learning agents in different settings and see that the hybrid agents consistently get a high win rate. The hybrid PPO agent performs the best in all scenarios with a win rate of 91\% against the baseline agents and a win rate of 69.06\% against other learning agents. In this work, we integrate a fixed-policy approach with a learning-based approach, but other hybrid approaches may be possible. For instance, instead of using a fixed-policy agent, the seldom occurring actions could be driven by a separate learning agent that could either be trained jointly or separately from the principal learning agent. In the future, we plan to explore other hybrid approaches, train multiple agents using Multi-Agent Reinforcement Learning (MARL) techniques, and extend the Monopoly simulator to support human opponents. 
\bibliographystyle{IEEEtran}
\bibliography{bibs/ref}
\clearpage
\pagenumbering{arabic}
\setcounter{page}{1}

\title{Supplementary for: Decision Making in Monopoly using a Hybrid Deep Reinforcement Learning Approach}
% note positions of commas and non-breaking spaces ( ~ ) LaTeX will not break
% \author{
% 	Trevor Bonjour$^{1*}$, Marina Haliem$^{1*}$, Aala Alsalem$^{1}$, Shilpa Thomas$^{2}$, Hongyu Li$^{2}$, \\ Vaneet Aggarwal$^{1}$, Mayank Kejriwal$^{2}$, and Bharat Bhargava$^{1}$
% 	\thanks{$^*$ Equal contribution $^1$ Purdue University, $^2$ University of Southern California	}
% 	}
	
\maketitle
\appendices
\color{black}

\section{Approach Details}\label{appendix_approach_details}
In the following sections, $t$ represents a time step in the game that the agent needs to make a decision. $s_t$ represents the state of the environment at time $t$ and is represented as a vector of size 240 as described in \Cref{sec_state}. $S$ represents all possible states of the environment. $a_t$ denotes the action that an agent takes at time $t$ and is represented as a 2922 dimensional vector (\Cref{sec_action}). $A$ denotes all possible actions. $A(s_t)$ denotes a set of all possible actions that are allowed in a given state $s_t$ at time $t$. $r_t$ denotes the reward the agent receives after taking an action $a_t$ at time $t$ and is represented as a scalar value which is calculated using \Cref{eq:overall_reward}. $R$ represents the set of all possible rewards. 
\begin{figure*}
    \centering
    \includegraphics[width = \textwidth]{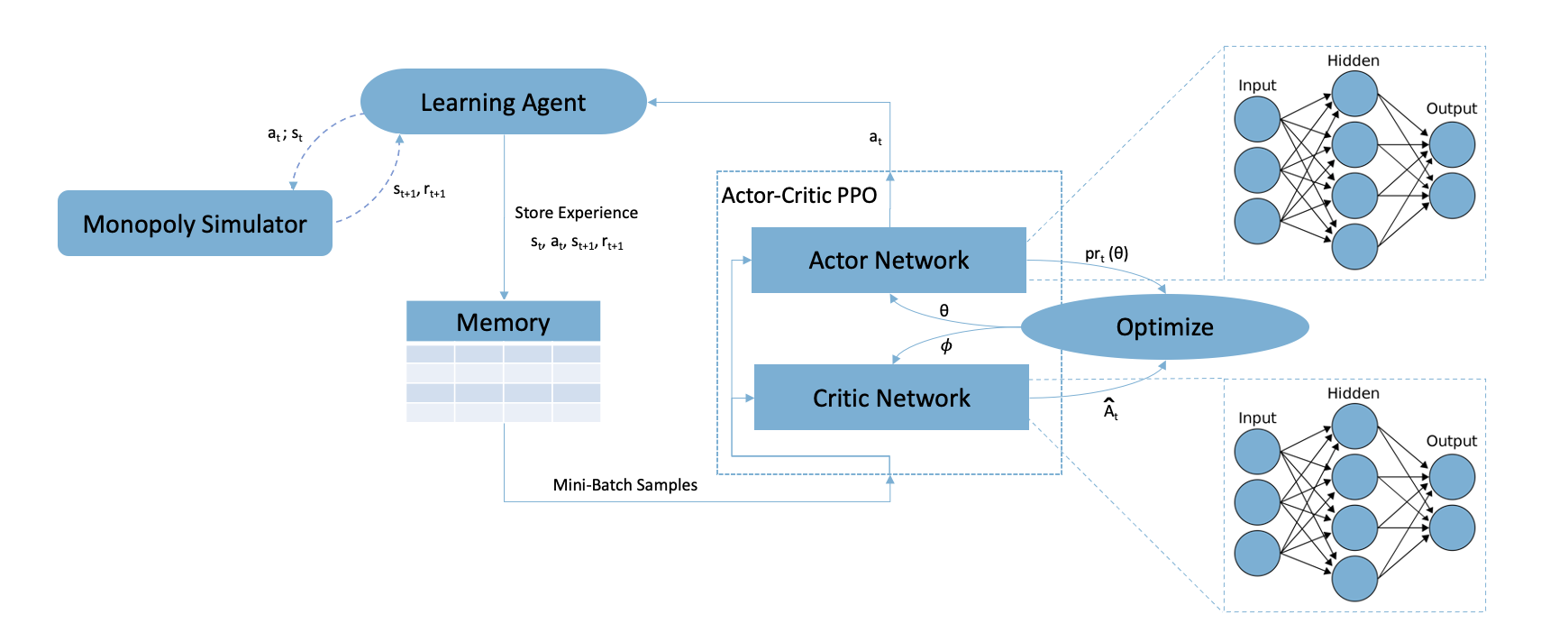}
    \caption{Actor-Critic PPO Approach for Monopoly}
    \label{ppo_arch}
\end{figure*}

\subsection{Actor-Critic PPO} \label{appendix_ppo_approach}
PPO is an on-policy algorithm in which the agent follows a policy dictated by the actor network for a fixed number of time-steps $N$ (much less than the episode length). The policy network parameters $\theta$ are initialized randomly, and the agent explores by sampling actions according to the latest version of its stochastic policy ($\pi_\theta$). As the training proceeds, the policy typically becomes less random since the update rule encourages it to exploit rewards. The critic network is responsible for calculating the value of a given state $s_t \in S$ at time $t$ given by $V_\phi(s_t)$ which is used to estimate the advantage $\hat{A_t}$. At each time-step $t$ the actor network returns an action $a_t \in A(s_{t})$ based on the current observed state $s_t \in S$ of the environment, where $S$ is the set of possible states and $A(s_{t})$ is the finite set of possible actions in state $s_{t}$. Once the action is executed the agent receives a reward $r_t \in R$ and the state of the environment is updated to $s_{t+1}$. The transitions of the form $<s_{t}, a_{t}, r_{t}, s_{t+1}>$ are stored in a memory along with the advantage estimates $\hat{A_t}$. The values $V_\phi(s_t)$ returned by the critic network are used in calculating the advantage estimates given by \Cref{eq_adv}.
\begin{multline}\label{eq_adv}
    \hat{A_t} = \delta_t + (\gamma\lambda)\delta_{t+1} + \dots + (\gamma\lambda)^{N-t+1}\delta_{N-1}
    \\ \text{where } \delta_t = r_t + \gamma V_\phi(s_{t+1}) - V_\phi(s_t)
\end{multline}
Once the memory is full, we randomly sample mini-batch transitions and optimize the PPO objective function (\Cref{ppo_obj}) and the mean squared error loss (\Cref{ppo_mse}) for the critic network. The procedure is given in \Cref{ppo_algo}. Each episode represents a complete game. Each time-step denotes every instance that the agent needs to take any action within the game.

\begin{multline}\label{ppo_obj}
\theta = \arg \max_{\theta} E [L(\theta)]\\
   \text{where } L(\theta) = \min\big( pr_t(\theta) \hat{A_t}, clip(pr_t(\theta), 1 - \epsilon, 1+\epsilon)\hat{A_t}\big)
   \\ \text{and } pr_t(\theta) = \frac{\pi_\theta(a_t|s_t)}{\pi_{\theta_{old}}(a_t|s_t)}
\end{multline}
$\pi_{\theta_{old}}$ denotes the policy before the update. 
\begin{equation}\label{ppo_mse}
\phi = \arg \min_{\phi} \big( V_\phi(s_t) - r_t\big)^2
\end{equation}

\begin{algorithm}[!htbp]
	\caption{\textcolor{black}{Actor-Critic PPO}}
	\label{ppo_algo}
	\begin{algorithmic}[1]
	\color{black}
		\State \textbf{Initialize} batch size $m$, trajectory memory $D$, memory size $N$, actor network parameters $\theta$ and critic network parameters $\phi$.
		\For{$e = 1:Episodes$}
		\State \textbf{Initialize} the game board with arbitrary order for player turns.
		\For{$t = 1:N$}
		\State \textbf{Get} current state $s_{t}$
		\State \textbf{Select} action $a_t$ by following policy $\pi_{\theta}$
		\State \textbf{Execute} action based on $a_{t}$
		\State \textbf{Calculate} reward $r_t$ and get new state $s_{t+1}$
		\State \textbf{Store} transition $(s_{t}, a_{t}, r_{t},s_{t+1})$ in $D$
		\State \textbf{Calculate} advantage estimates $\hat{A_t}$

		\EndFor
		\State \textbf{Sample} random batch of size $m$ from $D$.
		\State \textbf{Maximize} PPO objective w.r.t $\theta$
		\State \textbf{Minimize} MSE for value function w.r.t $\phi$
		\State Clear memory $D$ and proceed to step 4
		\EndFor
	\end{algorithmic}
\end{algorithm}
\color{black}
\subsection{DDQN}\label{appendix_ddqn_approach}

\begin{figure*}
    \centering
    \includegraphics[width = \textwidth]{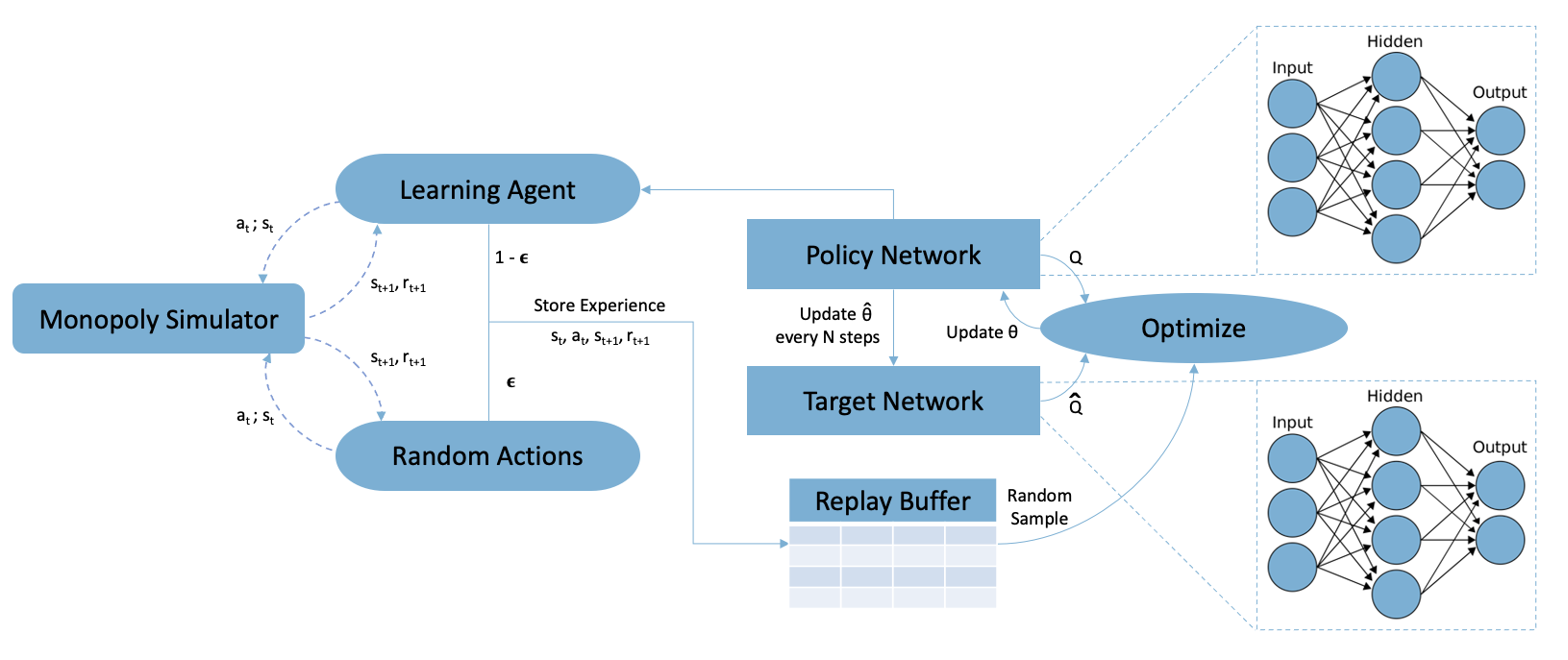}
    \caption{DDQN Approach for Monopoly}
    \label{arch}
\end{figure*}

\Cref{arch} shows the overall flow of our approach. At each time-step $t$, the DRL agent selects an action $a_{t} \in A(s_{t})$ based on the current state of the environment $s_{t} \in S$, where $S$ is the set of possible states and $A(s_{t})$ is the finite set of possible actions in state $s_{t}$. Similar to \cite{mnih2015human}, we make use of the $\epsilon$-greedy exploration policy to select actions. Initially, the agent explores the environment by randomly sampling from allowed actions. As the learning proceeds and the agent learns which actions are more successful, its exploration rate decreases in favor of more exploitation of what it has learned. We mask the output of the network to only the allowed actions to speed up the training process. The action masking ensures that the learning agent selects a valid action at any given time.

\color{black}
\begin{algorithm}[!htbp]
	\caption{Double Deep Q-learning with Experience Replay}
	\label{DQN_algo}
	\begin{algorithmic}[1]
		\State \textbf{Initialize} batch size $I$, replay buffer $D$, maximum memory $M$, policy Q-network parameters $\theta$ and target Q-network parameters $\hat{\theta}$.
		\For{$e = 1:Episodes$}
		\State \textbf{Initialize} the game board with arbitrary order for player turns.
		\State \textbf{Get} initial state $s_{0}$
		\For{$t = 1:T$}
		\State With probability $\epsilon$, \textbf{select} random action $a_t$ from valid actions
		\State \textbf{Else} $a_{t} \gets \text{argmax}_{a}Q(s_{t},a;\theta)$
		\State \textbf{Execute} action based on $a_{t}$
		\State \textbf{Calculate} reward $r_t$ and get new state $s_{t+1}$
		\State \textbf{Store} transition $(s_{t}, a_{t},r_{t},s_{t+1})$ in $D$
		\color{black}
		\If{len($D$) $> M$}
		\State \textbf{Sample} random batch of size $I$ from $D$.
		\For{$i = 1:I$}
		\State $y_i \leftarrow \gamma \hat{Q}(s_{i+1},\underset{a}{\text{argmax}} {Q}(s_{i+1},{a_{i}};\theta);\hat{\theta})$
		\State $z_{i} \leftarrow r_{i}+ y_i$
		\State \textbf{Minimize} $(z_{i}-Q(s_{i},a_{i};\theta))$ w.r.t. $\theta$.
		\State $\hat{\theta} \leftarrow \theta$ every $N$ steps.
		\EndFor
		\EndIf
		\color{black}
		\EndFor
		\EndFor
	\end{algorithmic}
\end{algorithm}

\textcolor{black}{After an action is executed, at time $t$ the agent receives a reward, $r_{t} \in R$, and state of the environment is updated to $s_{t+1}$. The reward $r_t$ is calculated for the DDQN agent as per \Cref{eq:overall_reward}.} The transitions of the form $<s_{t}, a_{t}, r_{t}, s_{t+1}>$ are stored in a cyclic buffer, known as the \textit{replay buffer}. This buffer enables the agent to train on prior observations by randomly sampling from them. We make use of a target network to calculate the temporal difference error. The target network parameters $\hat{\theta}$ are set to the policy network parameters $\theta$ every fixed number of steps. \Cref{DQN_algo} gives the procedure. Each episode represents a complete game. Each time-step denotes every instance that the agent needs to take any action within the game.

\section{Network Architecture and Parameters}\label{net_param}
\color{black}\subsection{Actor-Critic PPO Agents}\label{net_param_ppo} We use a fully connected feed-forward network for the actor and the critic networks. The input to the networks is the current state of the environment, $s_t$, represented as a 240-dimensional vector (\Cref{sec_state}). We make use of 2 hidden layers, that consist of 1024 and 512 neurons respectively, each with a rectified linear unit (ReLU) as the activation function:

\begin{equation}
    f(x) = \begin{cases}
    x\ \ \ \ \text{for}\ x\geq 0
    \\ 0 \ \ \ \ \text{otherwise}
    \end{cases}
\end{equation}

The output layer of the actor network has a dimension of $2922$, where each element represents the probability of taking an action. As discussed earlier, not all actions are valid at all times. We mask the output of the final layer to only the allowed actions. The output of the critic networks is a single scalar depicting the value of the current state. For training the actor network, we employ the Adam optimizer \cite{kingma2014adam}.   After tuning our network, we achieved the best results using the following parameters: $\gamma = 0.9999$ $\lambda = 0.95$, actor learning rate $\alpha_a = 10^{-6}$ , critic learning rate $\alpha_c = 10^{-6}$ batch size $ m = 5$, and a memory size $ N = 20$.
\color{black}
\subsection{DDQN Agents}\label{net_param_ddqn} We use a fully connected feed-forward network to approximate $Q(s_t,a_t)$  for the policy network. The input to the network is the current state of the environment, $s_t$, represented as a 240-dimensional vector (\Cref{sec_state}). We make use of 2 hidden layers, that consist of 1024 and 512 neurons respectively, each with a rectified linear unit (ReLU) as the activation function:

\begin{equation}
    f(x) = \begin{cases}
    x\ \ \ \ \text{for}\ x\geq 0
    \\ 0 \ \ \ \ \text{otherwise}
    \end{cases}
\end{equation}

The output layer has a dimension of $2922$, where each element represents the Q-value for each of the actions the agent can take. As discussed earlier, not all actions are valid at all times. We mask the output of the final layer to only the allowed actions. For training the network, we employ the Adam optimizer \cite{kingma2014adam} and use mean-square error as the loss function. We initialize the target network with the same architecture and parameters as the policy network. We update the parameter values of the target network to that of the policy network every 500 episodes and keep them constant otherwise. After tuning our network, we achieved the best results using the following parameters: $\gamma = 0.9999$ , learning rate $\alpha = 10^{-5}$ , batch size $ = 128$, and memory size $ = 10^4$.
\color{black}
\section{Additional Evaluation Results}\label{appendix_add_eval_results}
We present additional evaluation results for \textbf{Experiment 3 - Hybrid vs Standard vs Fixed-Policy} (\Cref{sec_results}). 

\begin{table}[!hbpt]
\color{black}
\caption{Results for hybrid and standard PPO agents against fixed-policy agents over five runs of 2000 games each}

\label{table_ppo_v34}
\centering
\begin{tabular}{ccccc}
\toprule 
Run & FP-A & FP-B & Std PPO & Hybrid PPO\\
\midrule
1 & 55 & 75 & 152 & 1718 \Bstrut \\
2 & 62 & 80 & 125 & 1733 \Bstrut \\
3 & 60 & 66 & 144 & 1730 \Bstrut \\
4 & 57 & 71 & 118 & 1754 \Bstrut \\
5 & 59 & 72 & 140 & 1729 \Bstrut \\
\midrule
Win Rate & 2.93\% & 3.64\% & 6.79\% & \textbf{86.64\%} \\
\bottomrule
\end{tabular}
\end{table}
\begin{table}[!hbpt]
\color{black}
\caption{Results for hybrid and standard PPO agents against fixed-policy agents over five runs of 2000 games each}
\label{table_ppo_v35}
\centering
\begin{tabular}{ccccc}
\toprule 
Run & FP-A & FP-C & Std PPO & Hybrid PPO\\
\midrule
1 & 45 & 67 & 116 & 1772 \Bstrut \\
2 & 43 & 52 & 128 & 1777 \Bstrut \\
3 & 48 & 57 & 118 & 1777 \Bstrut \\
4 & 42 & 57 & 115 & 1786 \Bstrut \\
5 & 37 & 57 & 147 & 1759 \Bstrut \\
\midrule
Win Rate & 2.15\% & 2.9\% & 6.24\% & \textbf{88.71\%} \\
\bottomrule
\end{tabular}
\end{table}
\begin{table}[!hbpt]
\caption{Results for hybrid and standard DDQN agents against fixed-policy agents over five runs of 2000 games each}

\label{table_v34}
\centering
\begin{tabular}{ccccc}
\toprule 
Run & FP-A & FP-B & Std DDQN & Hybrid DDQN\\
\midrule
1 & 229 & 345 & 272 & 1154 \Bstrut \\
2 & 226 & 297 & 299 & 1178 \Bstrut \\
3 & 235 & 305 & 236 & 1224 \Bstrut \\
4 & 258 & 342 & 276 & 1124 \Bstrut \\
5 & 253 & 303 & 269 & 1175 \Bstrut \\
\midrule
Win Rate & 12.01\% & 15.92\% & 13.52\% & \textbf{58.55\%} \\
\bottomrule
\end{tabular}
\end{table}
\begin{table}[!hbpt]

\caption{Results for hybrid and standard DDQN agents against fixed-policy agents over five runs of 2000 games each}
\label{table_v35}
\centering
\begin{tabular}{ccccc}
\toprule 
Run & FP-A & FP-C & Std DDQN & Hybrid DDQN\\
\midrule
1 & 190 & 283 & 310 & 1217 \Bstrut \\
2 & 206 & 286 & 304 & 1204 \Bstrut \\
3 & 235 & 269 & 316 & 1180 \Bstrut \\
4 & 207 & 282 & 318 & 1193 \Bstrut \\
5 & 218 & 293 & 299 & 1190 \Bstrut \\
\midrule
Win Rate & 10.56\% & 14.13\% & 15.47\% & \textbf{59.84\%} \\
\bottomrule
\end{tabular}
\end{table}

\begin{figure*}
    \centering
    \includegraphics[width = 0.8\textwidth]{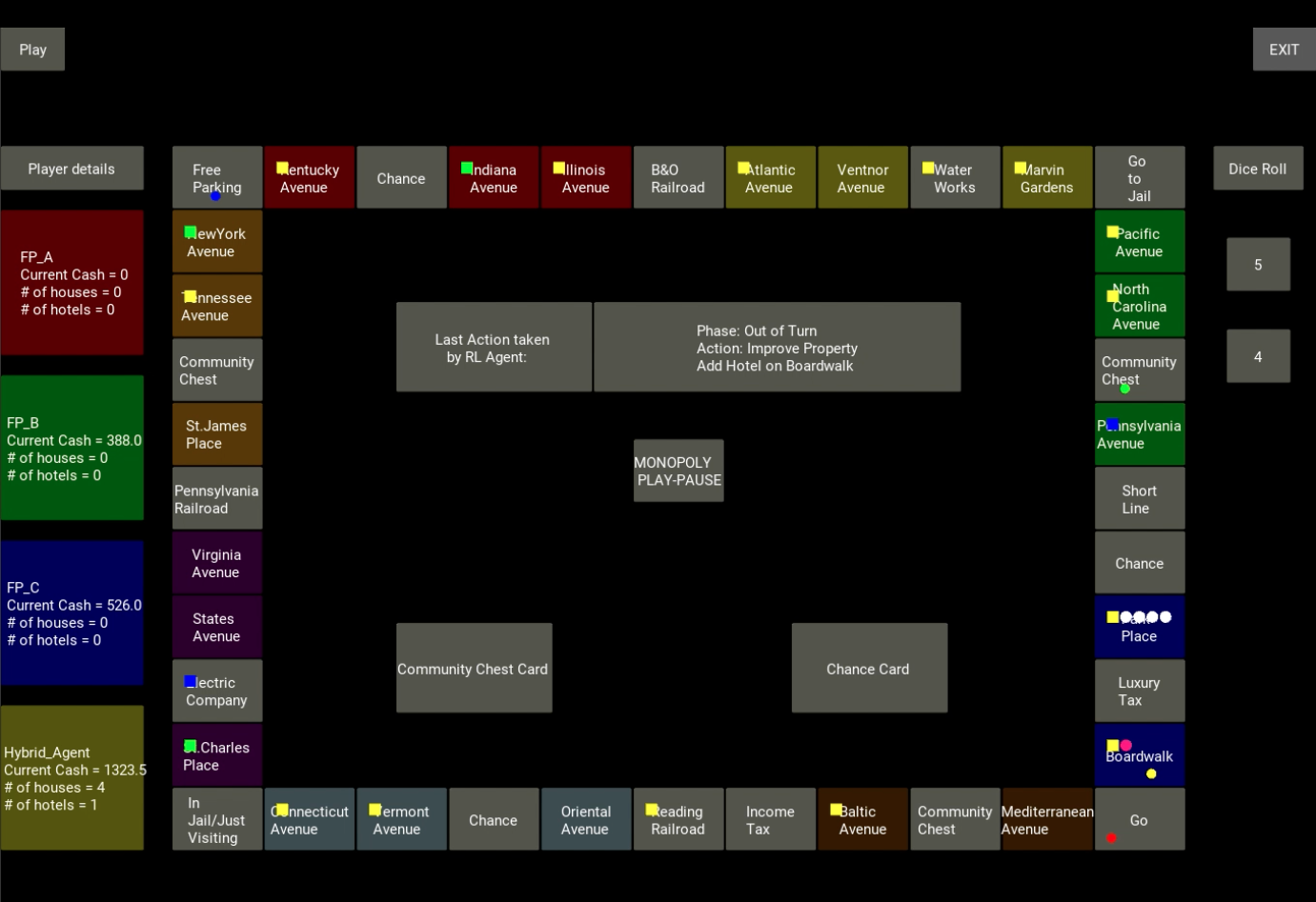}
    \caption{Screengrab of the gameplay video.}
    \label{vid_log}
\end{figure*}

\section{State Space with Embedded Memory}
\begin{figure}[!ht]
\centering
\includegraphics[width=0.48\textwidth]{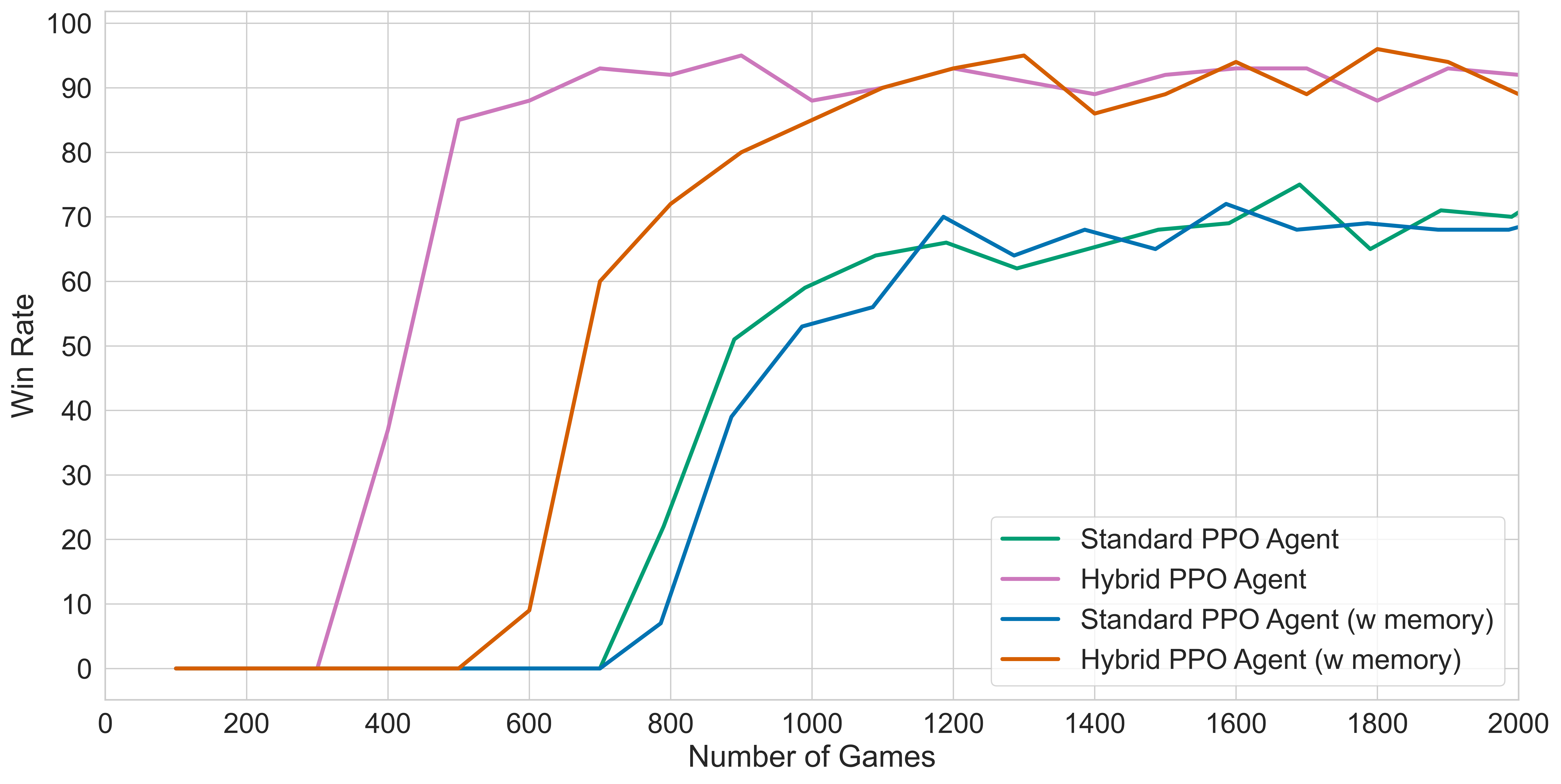}
\caption{Comparison of win rate (number of wins every 100 games) for PPO agents with and without using memory embedding in the state space.}
\label{fig_ppo_mem}
\end{figure}

We perform further experiments to see if the use of memory in the state representation would have an impact on the performance of our learning agents. We embed one-step look back into our state representation, where the agent memorizes the immediate previous state, hoping to improve the action choice for the current timestep. We show a comparison of the win rate of different PPO agents during training in \Cref{fig_ppo_mem}. We observe that adding the memory embedding to the state space gives similar performance but increases the convergence time.

\section{Visualization of Gameplay}
To visualize the decision-making process of the learning agents during the game, we create a visual simulator for Monopoly. We also create a detailed log of the gameplay in order to improve interpretability. We provide videos and logs of different games played by the hybrid agent as a supplement to the paper.

\Cref{vid_log} shows a screengrab from the simulation video for Monopoly. The blocks on the left show the player details. Each player is associated with a color and the order of the players on the left hand side shows the order in which they play the game. In \Cref{vid_log}, the order is FP-A $\rightarrow$ FP-B $\rightarrow$ FP-C $\rightarrow$ hybrid agent. Each player detail box consists of the current cash and the number of houses and hotels owned by the player. The square grid represents the Monopoly board (\Cref{fig_board}) with different properties. At the beginning of the game, each player starts at the bottom right corner which is the \emph{Go} position. \newpage

If a player goes bankrupt and is out of the game, their position is set to \emph{Go}. Each real-estate property is part of a color group as shown on the board. Each player and their current position is depicted on the square grid with a colored circle at the bottom of the cell. The color of each player is given on the left grid with player details. In \Cref{vid_log}, we see that FP-A is red (\tikz\draw[red,fill=red] (0,0) circle (.6ex);) and at the Go position since it is bankrupt, FP-B is green(\tikz\draw[green,fill=green] (0,0) circle (.6ex);), currently on the Community Chest cell, FP-C is blue(\tikz\draw[blue,fill=blue] (0,0) circle (.6ex);) currently at Free Parking cell and the hybrid agent is in yellow(\tikz\draw[yellow,fill={rgb:orange,1;yellow,2;pink,5}] (0,0) circle (.6ex);), currently on Boardwalk. The colored squares on the property cells represent the player that owns the respective property. For instance, in \Cref{vid_log}, the hybrid agent owns Vermont Avenue (\crule[{rgb:orange,1;yellow,2;pink,5}]{0.2cm}{0.2cm}), FP-B owns Charles Place (\crule[green]{0.2cm}{0.2cm}), and FP-C owns Electric Company(\crule[blue]{0.2cm}{0.2cm}). The white circle (\tikz\draw[black,fill=white] (0,0) circle (.6ex);) on the top of a cell represents a house and a pink circle (\tikz\draw[red!50!white!100,fill=red!50!white!100] (0,0) circle (.6ex);) represents a hotel. In \Cref{vid_log} we can see that the hybrid agent has 4 houses on Park Place and a hotel on Boardwalk. In order to show what action the learning agent takes, we provide the details of the last action taken along with the action parameters and the phase the action was taken in. These details are provided on the top rectangle in the center of the grid. In \Cref{vid_log}, we see that the hybrid agent chose the improve property action and added a hotel in the out of turn phase on Boardwalk.
\color{black}

% \square
% \crule{0.2cm}{0.2cm} test \crule[blue]{1cm}{1cm} \crule[red!50!white!100]{1cm}{1cm}  
% Some Text \tikz\draw[red,fill=red] (0,0) circle (.6ex); further text
% Some Text \tikz\draw[pink,fill=red!50!white!100] (0,0) circle (.6ex); further text
% Some Text \tikz\draw[black,fill=white] (0,0) circle (.6ex); further text

% ß\end{document}
% \clearpage
% \bibliography{bibs/ref}
\end{document}